\documentclass[letterpaper]{fairmeta}

\usepackage{amsmath,amsfonts,bm}

\def\eqref#1{equation~\ref{#1}}

\def\1{\bm{1}}

\DeclareMathAlphabet{\mathsfit}{\encodingdefault}{\sfdefault}{m}{sl}
\SetMathAlphabet{\mathsfit}{bold}{\encodingdefault}{\sfdefault}{bx}{n}

\usepackage{amsmath}
\usepackage{amssymb}
\usepackage{booktabs}
\usepackage{graphicx}
\usepackage{microtype}
\usepackage{multirow}
\usepackage{algorithm}
\usepackage{algpseudocode}
\usepackage{placeins}
\usepackage{array}
\usepackage{tikz}
\usetikzlibrary{arrows.meta,positioning}
\newcolumntype{P}[1]{>{\raggedright\arraybackslash}p{#1}}
\usepackage{xcolor}
\definecolor{peblue}{HTML}{235B8E}
\definecolor{peteal}{HTML}{247C79}
\definecolor{peamber}{HTML}{B67A28}
\usepackage{hyperref}

\usepackage{url}

\title{Discovered, Not Designed:\\
Population Evolution for Collaborative\\
and Compute-Intensive Model Discovery}
\author[*]{Bo Peng}
\author[*]{Lizhu Zhang}
\author{Yuhang Zhou}
\author{Mingyi Wang}
\author{Yifan Wu}
\author{Serena Li}
\author{Xiangjun Fan}
\author{Zhuokai Zhao}
\affiliation{Meta}
\contribution[*]{Equal contribution}
\hypersetup{
  pdftitle={Discovered, Not Designed: Population Evolution for Collaborative and Compute-Intensive Model Discovery},
  pdfauthor={Bo Peng, Lizhu Zhang, Yuhang Zhou, Mingyi Wang, Yifan Wu, Serena Li, Xiangjun Fan, Zhuokai Zhao},
  pdfsubject={Population Evolution and Recursive Model Discovery},
  pdfkeywords={}
}

\abstract{
LLM-driven evolution enables iterative model development, but two practical goals remain underexplored: finding model designs that transfer across related tasks and sustaining improvement when training is expensive.
We introduce \textbf{Population Evolution (PE)}, a collaborative, hierarchical framework that connects ongoing local searches through shared experimental evidence.
PE evaluates code changes across related training instances and shares the results to guide subsequent proposals and promotion to larger training scales.
For expensive targets, PE searches small training subsets and screens candidates through peer and intermediate evaluations before full-target training.
We introduce \textbf{RMD-Bench} to evaluate both settings across ranking, watch-time prediction, RL algorithm discovery, and LLM/VLM pretraining.
Compared with standalone evolution at matched source iterations, PE raises mean best local gains from 7.01\% to 8.97\% in ranking and from 2.84\% to 3.85\% in watch-time, while improving the best larger-scale outcome in all three joint-discovery families.
In watch-time discovery, PE improves best larger-scale gains with four of five harnesses and all four proposers.
On new recommendation datasets under shared target-side calibration, every evaluated PE design improves over the reference in mean performance.
Under matched total GPU compute, completed LLM discovery runs yield a best relative accuracy gain of 2.48\% and 13 successful candidates for PE, versus 0.92\% and none for direct evolution.
VLM loss reduction reaches 8.78\% versus 5.05\% under matched total GPU compute.
}

\date{October 4, 2026}
\correspondence{Bo Peng, Lizhu Zhang, and Zhuokai Zhao\\
\email{bopeng707@meta.com}, \email{lizhu@meta.com}, \email{zhuokai@meta.com}}

\begin{document}
\maketitle

\section{Introduction}
\label{sec:introduction}

LLM-driven evolution automates model development by proposing code changes, evaluating them, and using the results to guide further edits~\citep{liu2024eoh,novikov2025alphaevolve,lange2025shinkaevolve}.
Related research agents automate idea generation and experimentation~\citep{lu2024aiscientist}, while neural architecture discovery offers a route toward recursive self-improvement~\citep{pepe2026agentic}.
Despite this progress, two goals remain underexplored in LLM-driven neural model discovery: discovering universal model designs across related tasks and sustaining improvement when candidate training is expensive.

\textbf{Collaborative model discovery within a domain.}
Related datasets often share modeling requirements, so a mechanism discovered on one can inform development on another.
Independent single-task searches primarily use their own feedback~\citep{novikov2025alphaevolve,lange2025shinkaevolve}, leaving this evidence isolated.
Connecting these searches can support both local improvement and \emph{universal model designs}: a common architecture or training mechanism used across related instances, with parameters trained separately on each.

\textbf{Sustaining discovery on computationally demanding tasks.}
When every candidate requires substantial computation, as in full LLM or VLM training, a fixed budget permits few proposal--evaluation cycles.
Parallel full-target searches broaden exploration, but costly evaluations limit how many revisions each search can make~\citep{novikov2025alphaevolve}.
Sustained discovery therefore requires affordable experiments that inform continued development, followed by verification under full-target training.

We introduce \textbf{Population Evolution (PE)}, which connects ongoing local searches through a shared ledger of executable interventions and cross-instance evaluation results (Figure~\ref{fig:pe_overview}).

Each source retains its objective and archive; shared feedback guides local edits, while replay lets other instances reuse its discoveries.
Cross-instance replay tests the same intervention after retraining under each receiving instance's protocol.

In \textbf{joint discovery}, searches on related datasets or training configurations reuse, adapt, and combine mechanisms; replay and larger-scale evaluation establish where the resulting designs remain useful.
In \textbf{decomposed discovery}, small subsets of an expensive target's training data provide affordable search instances, with peer and intermediate evaluations directing candidates toward full-target training.
This hierarchy supports iterative refinement while reserving full-target training for screened candidates.

We introduce the \textbf{Recursive Model Discovery Benchmark (RMD-Bench)}, providing executable evaluation protocols for five task families spanning both settings.
Compatible interfaces and fixed references support evaluation of local gains and transfer in joint discovery, and full-target quality and yield in decomposed discovery.
All three joint-discovery families also test transfer to new datasets or disjoint coding tasks.
We start from FuXi-Linear~\citep{ye2026fuxilinear} for ranking, EGMN~\citep{zhao2025egmn} for watch-time prediction, GRPO~\citep{shao2024deepseekmath} for RL algorithm discovery, Gemma~3-style architectures~\citep{gemmateam2025gemma3} for LLM pretraining, and Qwen3.5-style architectures~\citep{qwenteam2026qwen35} for VLM pretraining.
A complete LLM or VLM discovery run, including five searches and their evaluations, requires thousands of GPU-hours.

At matched source iterations, PE improves the mean per-source best gain in both recommendation families and the best larger-scale outcome in all three joint-discovery families.
Every evaluated PE recommendation design improves over the reference mean on new datasets.
Retraining uses shared target-side calibration without candidate-specific tuning.
For LLM pretraining, completed runs under matched total GPU compute yield a 2.48\% best relative accuracy gain and 13 successful candidates for PE, versus 0.92\% and none for direct evolution.
For VLM pretraining, best loss reduction rises from 5.05\% to 8.78\% under matched total GPU compute.
Our contributions are:
\begin{itemize}
\item \textbf{Population Evolution:} a collaborative and hierarchical framework that uses shared experimental feedback to guide local search, reuses executable interventions across instances, and promotes candidates across training scales.
\item \textbf{RMD-Bench:} five executable task families with compatible code interfaces, shared reference implementations, and controlled cross-instance and cross-scale evaluation, including transfer to new datasets and tasks in all three Track~A families.
\item \textbf{Model discoveries and transfer:} recommendation designs that improve over shared references on new datasets, better larger-scale outcomes in all three Track~A families, and successful full-target LLM/VLM interventions, with watch-time comparisons across harnesses and proposers.
\end{itemize}

\begin{figure}[!t]
\centering
\resizebox{\linewidth}{!}{%
\begin{tikzpicture}[y=.70cm,
 font=\fontfamily{phv}\fontsize{8}{9.5}\selectfont,text=black!90,
 box/.style={draw=black!65,line width=.6pt,fill=white,align=center,inner sep=2.5pt,minimum height=.60cm},
 source/.style={box,draw=peblue!80,fill=peblue!5},
 evidence/.style={box,draw=black!65,fill=black!3},
 flow/.style={-{Latex[length=1.5mm]},draw=black!75,line width=.65pt},
 feedback/.style={-{Latex[length=1.35mm]},draw=black!60,dashed,line width=.6pt},
 head/.style={anchor=west,font=\fontfamily{phv}\fontsize{9}{10.5}\selectfont\bfseries},
 subhead/.style={anchor=west,font=\fontfamily{phv}\fontsize{8}{9.5}\selectfont\bfseries},
 note/.style={font=\fontfamily{phv}\fontsize{7}{8.2}\selectfont,text=black!70},
 lab/.style={note,fill=white,inner sep=1.3pt}
]
\node[head] at (0,.22) {(a) Ongoing local search and cross-instance replay};
\node[source,minimum width=3.2cm,minimum height=1.4cm] (local) at (1.7,-1.8) {\textbf{Local search on $E_i$}\\[2pt]Propose $h_{i,t}$ with $p_i$\\[-1pt]Train and evaluate\\[-1pt]Local history $\tau_{i,t}$};
\draw[feedback] (local.north east)--++(0,.3)-|(local.north west);
\node[lab,anchor=south] at (1.7,-.46) {local feedback};
\node[evidence,minimum width=3.25cm,minimum height=1.4cm] (mem) at (6.05,-1.8) {\textbf{Shared ledger $M_t$}\\[2pt]Executable interventions\\[-1pt]and evaluation results};
\draw[flow] (local.east)--node[lab,above=2pt]{publish}(mem.west);
\draw[draw=black!65,fill=black!2,line width=.6pt] (9.1,-.52) rectangle (14.4,-5.10);
\node[font=\fontfamily{phv}\fontsize{9}{10.5}\selectfont\bfseries] at (11.75,-.92) {Cross-instance replay on peer $E_j$};
\node[box,minimum width=4.8cm,minimum height=.60cm] (restore) at (11.75,-1.8) {Restore $E_j$'s reference code};
\node[box,minimum width=4.8cm,minimum height=.60cm] (apply) at (11.75,-2.82) {Apply the same published intervention};
\node[box,minimum width=4.8cm,minimum height=.75cm] (evaluate) at (11.75,-3.95) {Train and evaluate on $E_j$\\[-1pt]Use $E_j$'s prescribed protocol};
\draw[flow] (mem.east)--node[lab,above=2pt,xshift=-.12cm]{intervention}(restore.west);
\draw[flow] (restore.south)--(apply.north);
\draw[flow] (apply.south)--(evaluate.north);
\node[note] at (11.75,-4.80) {Repeat on peers; train weights separately};
\draw[feedback] ([xshift=-1.1cm]mem.south)--(4.95,-3.25)--(1.7,-3.25)--(local.south);
\node[note,align=center] at (3.25,-3.67) {Shared context $\Phi_i(M_t)$\\Guides the next proposal};
\draw[feedback] (evaluate.west)--(6.05,-3.95)--(mem.south);
\node[note] at (7.60,-4.29) {All peer results};
\node[note,align=center] at (4.40,-4.79) {Later proposals can reuse, modify, or combine peer mechanisms};
\node[note] at (7.2,-5.42) {Every source runs this cycle; peer evidence also guides promotion in (b) and (c)};
\draw[black!25,line width=.45pt] (0,-5.78)--(14.4,-5.78);
\begin{scope}[shift={(0,-.88)}]
\node[head] at (0,-5.58) {(b) Joint discovery (Track A)};
\node[head] at (7.6,-5.58) {(c) Decomposed discovery (Track B)};
\node[note] at (3.55,-6.03) {Related datasets or training configurations (five sources)};
\node[note] at (10.85,-6.03) {Training data at each scale (percent of target corpus)};
\foreach \x/\name in {1.15/1,3.55/2,5.95/5} {
 \node[source,minimum width=1.95cm,font=\fontfamily{phv}\fontsize{8}{9.5}\selectfont] (bsource\name) at (\x,-6.7) {Source\\[-1pt]instance $E_{\name}$};
}
\node[note,inner sep=0pt] at (4.75,-6.7) {$\cdots$};
\foreach \x/\name/\pct in {8.25/D2-A/2,9.55/D2-B/2,10.85/D5-A/5,12.15/D5-B/5,13.45/D5-C/5} {
 \node[source,minimum width=1.06cm,font=\fontfamily{phv}\fontsize{7.5}{8.6}\selectfont] (csource\name) at (\x,-6.7) {\name\\[-1pt]\pct\%};
}
\node[evidence,minimum width=6.2cm,minimum height=.50cm] (bm) at (3.55,-7.85) {Shared ledger};
\node[evidence,minimum width=5.6cm,minimum height=.50cm] (cm) at (10.85,-7.85) {Shared ledger};
\foreach \id in {1,2,5} {
 \draw[flow] ([xshift=-2pt]bsource\id.south)--([xshift=-2pt]bsource\id.south |- bm.north);
 \draw[feedback] ([xshift=2pt]bsource\id.south |- bm.north)--([xshift=2pt]bsource\id.south);
}
\foreach \id in {D2-A,D2-B,D5-A,D5-B,D5-C} {
 \draw[flow] ([xshift=-2pt]csource\id.south)--([xshift=-2pt]csource\id.south |- cm.north);
 \draw[feedback] ([xshift=2pt]csource\id.south |- cm.north)--([xshift=2pt]csource\id.south);
}
\node[box,draw=black!65,fill=black!2,minimum width=4.8cm,minimum height=.66cm] (bs) at (3.55,-8.95) {Cross-instance replay\\[-1pt]{\fontsize{7}{8.2}\selectfont Select candidates using peer results}};
\node[box,draw=black!65,fill=black!2,minimum width=4.8cm,minimum height=.66cm] (cs) at (10.85,-8.95) {Cross-instance replay\\[-1pt]{\fontsize{7}{8.2}\selectfont Select candidates using peer results}};
\draw[flow] (bm.south)--(bs.north);
\draw[flow] (cm.south)--(cs.north);
\node[box,minimum width=4.8cm] (bl) at (3.55,-10.65) {Larger-scale evaluation};
\draw[flow] (bs.south)--node[lab,right=3pt]{promote}(bl.north);
\node[box,minimum width=4.8cm,minimum height=.74cm] (bo) at (3.55,-12.40) {Universal model designs\\[-1pt]{\fontsize{7}{8.2}\selectfont Across related instances and training scales}};
\draw[flow] (bl.south)--(bo.north);
\draw[feedback] (bl.east)--(6.95,-10.65)--(6.95,-7.85)--(bm.east);
\draw[black!60,dashed,line width=.6pt] (bs.east)--(6.95,-8.95);
\node[box,minimum width=2.45cm,minimum height=.74cm,font=\fontfamily{phv}\fontsize{7}{8.2}\selectfont] (ca) at (9.45,-10.65) {Disjoint data\\[-1pt]D10 clean: $\approx10\%$};
\node[box,minimum width=2.45cm,minimum height=.74cm,font=\fontfamily{phv}\fontsize{7}{8.2}\selectfont] (cb) at (12.25,-10.65) {Expanded D5-C\\[-1pt]D10+C: $\approx10\%$};
\draw[flow] (cs.south)--(10.85,-9.83)-|(ca.north);
\draw[flow] (cs.south)--(10.85,-9.83)-|(cb.north);
\draw[black!75,line width=.65pt] (ca.south)--(9.45,-11.55)--(12.25,-11.55)--(cb.south);
\node[box,minimum width=5.05cm,minimum height=.74cm] (cf) at (10.85,-12.40) {Full-target training and evaluation: 100\%};
\draw[flow] (10.85,-11.55)--node[lab,right=3pt]{both pass}(cf.north);
\draw[feedback] (ca.west)--(7.78,-10.65)--(7.78,-7.85)--(cm.west);
\draw[feedback] (cb.east)--(14.15,-10.65)--(14.15,-7.85)--(cm.east);
\draw[black!60,dashed,line width=.6pt] (cs.east)--(14.15,-8.95);
\draw[flow] (.65,-13.55)--(1.30,-13.55);
\node[note,anchor=west] at (1.41,-13.55) {Executable interventions / evaluation};
\draw[feedback] (7.6,-13.55)--(8.25,-13.55);
\node[note,anchor=west] at (8.36,-13.55) {Evaluation evidence / proposal context};
\end{scope}
\end{tikzpicture}

}
\caption{\textbf{PE connects local searches through replay evidence and cross-scale promotion.}
(a) Published interventions are retrained on peers; their results enter the shared ledger and inform later proposals (Equation~\ref{eq:proposal-policy}).
(b) Track~A tests the same designs across related instances and at larger training scales.
(c) Decomposed discovery screens candidates on small sources and both intermediate instances before full-target training.
In Track~B, full-target results remain outside proposal context.}
\label{fig:pe_overview}
\end{figure}

\section{Population Evolution}
\label{sec:method}

An instance $E_i$ specifies the data, objective, and training/evaluation protocol.
In standard single-task evolution, an LLM repeatedly edits model or training code on one instance, using evaluation feedback to guide subsequent proposals~\citep{liu2024eoh,novikov2025alphaevolve}; we call this a \emph{local search}.
PE connects local searches across related \emph{source instances} $\mathcal D=\{E_i\}$, sharing evidence to guide proposals and promotion to larger training scales.
A \emph{candidate} is an executable intervention in model architecture or training code.
A \emph{shared ledger} $M_t$ stores candidate records and evaluation results.
Figure~\ref{fig:pe_overview} summarizes local search, cross-instance sharing, and promotion; Algorithm~\ref{alg:pe} in Appendix~\ref{app:algorithm} gives the asynchronous loop.

\subsection{Local search on each source instance}
\label{sec:lineages}
Each local search selects a parent from its own code archive. An LLM edits that parent's model or training code, and the resulting candidate is trained and evaluated on the same source instance.
Evaluation updates the local history and archive; PE supplements this feedback with shared experimental evidence.
At proposal step $t$, proposer $p_i$ generates candidate $h_{i,t}$ from local history $\tau_{i,t}$ and shared evidence $\Phi_i(M_t)$:
\begin{equation}
h_{i,t}\sim p_i\!\left(h\mid\tau_{i,t},\Phi_i(M_t)\right).
\label{eq:proposal-policy}
\end{equation}
With the local harness, proposer, and archive-update rules fixed, omitting $\Phi_i(M_t)$ reduces the source-search procedure to standalone evolution.
Candidate code is inherited and edited; model parameters are trained separately for each evaluation.

\subsection{Sharing and evaluating candidates across instances}
\label{sec:ideas}\label{sec:transfer}\label{sec:memory}
When a candidate meets its source's configured improvement criteria, PE publishes its executable intervention, mechanism description, provenance, and score to the shared ledger.
The other source instances are its \emph{peers}. PE restores each peer's reference code, applies the published intervention, then trains and evaluates it under that peer's protocol. We call this \emph{cross-instance replay} (Figure~\ref{fig:pe_overview}(a)).
All peer results, including weak or unsuccessful transfers, are added to the ledger.

Before every LLM call, $\Phi_i(M_t)$ summarizes the instance's own published candidates and peer transfers to $E_i$, including mechanisms and measured outcomes of successful and unsuccessful transfers.
Peer and promotion results affect local proposals only through this context; code archives remain instance-specific.
The proposer can use these outcomes to reuse, modify, or combine mechanisms.
Appendices~\ref{app:memory-projection} and~\ref{app:agent-prompts} detail the shared records, proposal context, and prompt structure.

\subsection{Promoting candidates to larger training scales}
\label{sec:selection}\label{sec:execution}
Shared results also govern promotion: published candidates must meet the configured improvement criteria on at least two of their four peers in our experiments (Appendix~\ref{app:selection-thresholds}).

In \textbf{joint discovery (Track A)}, eligible candidates are retrained on a larger-scale instance. Peer and larger-scale results return to the ledger, supporting local improvement and identifying designs useful across related instances (Figure~\ref{fig:pe_overview}(b)).

In \textbf{decomposed discovery (Track B)}, source searches run on affordable training subsets. Eligible candidates are tested on two intermediate instances; those passing both advance to full-target training (Figure~\ref{fig:pe_overview}(c)). Peer and intermediate results inform further proposals, while full-target scores and pass/fail outcomes are withheld from proposal context.
This is intended to prevent a few full-target successes from prematurely steering all source searches toward similar designs.
A candidate \emph{graduates} when it passes the final stage: larger-scale evaluation in Track~A or full-target evaluation in Track~B.
Local searches continue while these evaluations run asynchronously.

\section{RMD-Bench: Tasks and Evaluation}
\label{sec:benchmark}\label{sec:setting}

RMD-Bench defines five task families with objectives, reference implementations, editable components, and executable training/evaluation protocols.
Compatible interfaces support retraining across related instances. The 34 core instances vary data, training seed, or scale; Track~A also evaluates transfer beyond discovery instances.
The benchmark fixes evaluation conditions within each instance; each discovery system chooses its search, sharing, and selection policies.
Table~\ref{tab:benchmark_comparison} compares benchmark support for these settings; Section~\ref{sec:related-work} discusses related work, and Appendix~\ref{app:datasets} gives task specifications.

\subsection{Joint discovery (Track A)}
\label{sec:task-family}
Track~A tests local improvement and discovery of universal model designs useful across related tasks.

\textbf{Discovery instances.}\quad
Ranking, watch-time prediction, and RL algorithm discovery each have five source instances and one larger-scale instance (18 instances in total; Figure~\ref{fig:pe_overview}(b)).
Within each recommendation family, the five sources use different datasets; watch-time includes CIKM16 and WeChat. Both families use KuaiRand-27K for larger-scale evaluation.
Ranking edits sequence representations and temporal interactions in FuXi-Linear~\citep{ye2026fuxilinear}, using the mean of NDCG@10, HR@10, and MRR as its score.
Watch-time discovery edits EGMN's distributional head and point-prediction rule~\citep{zhao2025egmn}, using MAE in seconds.
RL discovery edits the GRPO loss~\citep{shao2024deepseekmath} and rollout filtering for Qwen2.5-Coder policies~\citep{hui2024qwen25coder}.
The five RL sources train 1.5B policies on the same task with different seeds; the larger-scale instance trains a 3B policy and evaluates pass@1 on a 423-problem LiveCodeBench partition~\citep{jain2024livecodebench}.
Appendices~\ref{app:ranking}--\ref{app:rl} specify the three protocols.

\textbf{Transfer beyond discovery instances.}\quad
\label{sec:benchmark-generalization}
Track~A evaluates designs beyond the discovery instances: KuaiVideo for ranking, VK-LSVD for watch-time prediction, and 632 LiveCodeBench tasks disjoint from the larger-scale selection partition for RL.
The two recommendation datasets are excluded from source search, peer evaluation, and promotion. KuaiVideo also changes the objective from next-item ranking to candidate-conditioned click-through rate (CTR) prediction.
Recommendation retraining uses a shared configuration calibrated and scored on each target's released test split, without candidate-specific redesign or tuning. RL evaluation uses held-out coding tasks.
These evaluations supplement the 34 core instances; Appendix~\ref{app:postdiscovery-transfer} specifies selection, calibration, and reporting.

\subsection{Decomposed discovery (Track B)}
\label{sec:heavy-tasks}
Track~B evaluates whether discovery on small training subsets produces effective model and training designs for an expensive full target within a compute budget.

\textbf{Training hierarchy.}\quad
LLM and VLM pretraining each have five source instances, two intermediate instances, and one full-target instance (16 instances in total; Figure~\ref{fig:pe_overview}(c)).
The hierarchy tests transfer across training subsets and training scales before committing to expensive full-target evaluation.
The sources include three disjoint 5\% subsets, D5-A/B/C. D2-A and D2-B each contain 2\% of the target corpus and are nested in D5-A and D5-B, respectively.
Disjoint peers test whether gains persist beyond the source subset; the nested 2\%/5\% instances test transfer across data volumes and numbers of training steps.
Two approximately 10\% intermediate instances test different forms of transfer: D10 clean uses data disjoint from every source, while D10+C combines D5-C with a fresh 5\% subset.
Requiring success on both is intended to prioritize designs whose gains persist under full-target training.
Full-target evaluation uses 100\% of the training corpus. The same executable intervention is evaluated at each tier under its prescribed training configuration.

\textbf{Pretraining tasks.}\quad
LLM discovery starts from a 185M-parameter Gemma~3-style decoder~\citep{gemmateam2025gemma3} and can modify its architecture, optimizer, and training mechanisms.
Training uses the FineWeb-Edu 100B core set~\citep{lozhkov2024finewebedu,karpathy2025finewebedu100b}.
The score is mean accuracy over the seven zero-shot tasks used in Mamba-2~\citep{dao2024mamba2}: ARC-Challenge, ARC-Easy, HellaSwag, LAMBADA OpenAI, OpenBookQA, PIQA, and WinoGrande.
VLM discovery edits a Qwen3.5-style backbone~\citep{qwenteam2026qwen35}, projector, optimizer, and schedule, retaining a frozen SigLIP encoder~\citep{zhai2023siglip}.
The starting VLM has approximately 646M trainable parameters; full-target training uses 2.38M raw FineVisionMax-format examples~\citep{wiedmann2025finevision}. The score is token-weighted validation cross-entropy.
Appendices~\ref{app:llm} and~\ref{app:vlm} give the full specifications.

\subsection{Evaluation and reporting}
\label{sec:interfaces}
Each intervention is evaluated by training a model under the receiving instance's prescribed initialization, data, and training allocation.
For candidate score $y_e(h)$ and fixed reference score $b_e$ on instance $e$, relative improvement is
\begin{equation}
\label{eq:rmd-gain}
r_e(h)=\frac{s_e\bigl(y_e(h)-b_e\bigr)}{\max(|b_e|,\epsilon)},
\end{equation}
where $s_e=+1$ for ranking score or accuracy and $-1$ for MAE or loss; $\epsilon>0$ prevents division by zero. Reported percentage gains are $100r_e(h)$.
Track~A reports local gains, larger-scale transfer, and transfer beyond discovery instances. Track~B reports full-target quality and discovery yield: the number of candidates meeting predefined full-training success criteria.
Section~\ref{sec:experiments} specifies the comparisons; Appendices~\ref{app:baseline-construction}, \ref{app:benchmark-runtime}, and~\ref{app:execution-accounting} describe baseline calibration, discovery compute scales, and budget accounting.

\begin{table}[!ht]
\centering
\fontsize{8}{9}\selectfont
\setlength{\tabcolsep}{2pt}
\renewcommand{\arraystretch}{1.0}
\caption{\textbf{Explicit benchmark support for collaborative and compute-intensive discovery.}}
\label{tab:benchmark_comparison}
\begin{tabular*}{\linewidth}{@{\extracolsep{\fill}}P{3.55cm}*{7}{>{\centering\arraybackslash}p{1.30cm}}@{}}
\toprule
\textbf{Evaluation support} & \shortstack{MLE-\\bench\\\citeyearpar{chan2024mlebench}} & \shortstack{Innovator\\Bench\\\citeyearpar{wu2025innovatorbench}} & \shortstack{Research\\Gym\\\citeyearpar{garikaparthi2026researchgym}} & \shortstack{PostTrain\\Bench\\\citeyearpar{rank2026posttrainbench}} & \shortstack{MLS-\\Bench\\\citeyearpar{lyu2026mlsbench}} & \shortstack{Edge\\Bench\\\citeyearpar{zhu2026edgebench}} & \shortstack{\textbf{RMD-}\\\textbf{Bench}\\\textbf{(ours)}} \\
\midrule
Collaboration protocol & --- & --- & --- & --- & --- & --- & $\checkmark$ \\
Cross-dataset transfer & --- & --- & $\checkmark$ & --- & $\checkmark$ & --- & $\checkmark$ \\
Cross-scale transfer & --- & --- & --- & --- & $\checkmark$ & --- & $\checkmark$ \\
\midrule
Hierarchical training & --- & --- & --- & --- & $\triangle$ & --- & $\checkmark$ \\
Discovery yield vs.\ compute & --- & --- & --- & --- & --- & --- & $\checkmark$ \\
\midrule
Calibrated starting configurations & --- & --- & $\checkmark$ & --- & $\checkmark$ & --- & $\checkmark$ \\
\bottomrule
\end{tabular*}
\par\vspace{1pt}
\begin{minipage}{\linewidth}\fontsize{8}{9}\selectfont
$\checkmark$: explicit support; ---: not specified; $\triangle$: partial support in MLS-Bench's adaptive pretraining study.
Collaboration protocols support exchanging discoveries during ongoing search; transfer evaluates a common intervention across datasets or training scales.
Discovery yield counts candidates meeting predefined full-training success criteria; Track~B compares completed runs under matched total GPU compute.
\end{minipage}
\end{table}

\section{Experiments}
\label{sec:experiments}

Experiments test local discovery and transfer in Track~A, and full-target quality and completed-run yield in Track~B. Section~\ref{sec:analysis} analyzes peer evaluation and shared evidence.

\subsection{Experimental setup and comparisons}
\label{sec:pe_setup_and_reporting}
\textbf{Shared protocols.}\quad
The main experiments use the AdaEvolve-based search harness~\citep{cemri2026adaevolve} with GPT-5.5 at high reasoning effort as the proposer (Appendix~\ref{app:implementation}).
Methods share starting models, reference configurations, editable interfaces, instructions, and training/evaluation protocols (Appendix~\ref{app:environment-preparation}). Gains use fixed instance references (Equation~\ref{eq:rmd-gain}).
Section~\ref{sec:pe_applicability_across_harnesses_and_proposal_models} varies the harness and proposer in watch-time discovery.

\textbf{Track A: independent and collaborative discovery.}\quad
Standalone evolution (S) runs independent source searches using the same harness and parent-selection rules as PE, without shared context. Both methods run five 100-iteration searches. Table~\ref{tab:pe_joint}A--C compares locally proposed candidates with and without shared feedback; peer-replayed candidates are reported separately (Section~\ref{sec:transfer-analysis}).
Local recommendation results average the best gain from each dataset; RL uses the best result across five searches of the same task.
At larger scale, S retrains each source's top candidates by source performance, while PE retrains promoted candidates. S's per-source quota is rounded up to give it at least as many larger-scale evaluations as PE (Table~\ref{tab:pe_joint}D).
Neither method searches locally at larger scale.
Transfer evaluations use shared target protocols without candidate-specific redesign or tuning (Appendix~\ref{app:transfer-protocol}).

\textbf{Track B: hierarchical discovery under a compute budget.}\quad
PE screens candidates from five subset searches at intermediate scale before full training; direct evolution (Direct) runs five full-target searches. Completed runs use matched total GPU compute within each family. PE proposals use only source and intermediate feedback; Direct uses full-target feedback (Appendix~\ref{app:execution-accounting}).
\subsection{Joint discovery improves local quality and generalization}
\label{sec:joint-results}

We examine local gains, transfer within the discovery hierarchy, and transfer to new datasets or tasks.
\label{sec:pe_track_a_joint_discovery_across_related_instances}

\textbf{Shared feedback improves mean best local recommendation gains.}\quad
At the matched source allocation, PE raises mean best local gains from 7.01\% to 8.97\% in ranking and 2.84\% to 3.85\% in watch-time, exceeding S on 3/5 and 4/5 datasets, respectively (Table~\ref{tab:pe_joint}).
PE's mean gains over calibrated starting models are non-trivial in magnitude, comparable to published improvements over the strongest reported baselines on public datasets: 0.71--12.15\% in NDCG@10 for FuXi-Linear~\citep{ye2026fuxilinear} and 1.72--5.04\% in MAE for EGMN~\citep{zhao2025egmn}.
The best RL source gain rises from 23.53\% to 32.35\% (Table~\ref{tab:pe_joint}C).

\textbf{Selected designs improve larger-scale outcomes.}\quad
After larger-scale retraining without local search, PE yields more successful evaluations than S and raises best gains from 4.70\% to 6.19\% in ranking, 1.77\% to 2.17\% in watch-time, and 1.95\% to 6.64\% in RL (Table~\ref{tab:pe_joint}D).
These larger-scale instances participate in discovery; we next evaluate transfer to new datasets and tasks.

\begin{table}[!ht]
\centering\fontsize{8}{9}\selectfont
\setlength{\tabcolsep}{3pt}
\renewcommand{\arraystretch}{1.0}
\caption{\textbf{PE improves mean best local recommendation gains and best larger-scale outcomes.}
A--C report relative gains (\%) from locally proposed candidates at matched source iterations; C also gives solved-task counts. D reports successful/total larger-scale evaluations and best gains after retraining without local search.}
\label{tab:pe_joint}
\begin{minipage}[t]{.485\linewidth}
\centering
\textbf{A. Ranking: best source gain}\par\vspace{1pt}
\begin{tabular*}{\linewidth}{@{\extracolsep{\fill}}lrr@{}}
\toprule
Dataset & S & PE \\
\midrule
KuaiLive & 8.93 & \textbf{23.19} \\
KuaiRand-1K & 5.53 & \textbf{9.29} \\
KuaiRand-Pure & \textbf{8.03} & 6.07 \\
KuaiRec & 1.03 & \textbf{2.42} \\
KuaiSAR & \textbf{11.52} & 3.88 \\
\midrule
Mean & 7.01 & \textbf{8.97} \\
\bottomrule
\end{tabular*}
\end{minipage}\hfill
\begin{minipage}[t]{.485\linewidth}
\centering
\textbf{B. Watch-time: best source gain}\par\vspace{1pt}
\begin{tabular*}{\linewidth}{@{\extracolsep{\fill}}lrr@{}}
\toprule
Dataset & S & PE \\
\midrule
CIKM16 & 7.32 & \textbf{12.55} \\
KuaiRec & 0.40 & \textbf{0.67} \\
WeChat & 0.80 & \textbf{0.81} \\
KuaiRand-Pure & \textbf{0.75} & 0.16 \\
KuaiRand-1K & 4.93 & \textbf{5.08} \\
\midrule
Mean & 2.84 & \textbf{3.85} \\
\bottomrule
\end{tabular*}
\end{minipage}
\par\vspace{1pt}
\begin{minipage}[t]{.32\linewidth}
\centering
\textbf{C. RL: best source outcome}\par\vspace{1pt}
\fontsize{8}{9}\selectfont
\setlength{\tabcolsep}{2pt}
\begin{tabular*}{\linewidth}{@{\extracolsep{\fill}}P{.48\linewidth}rr@{}}
\toprule
Summary & S & PE \\
\midrule
Best gain (\%) & 23.53 & \textbf{32.35} \\
Solved / 408 & 14 & \textbf{15} \\
\bottomrule
\end{tabular*}
\end{minipage}\hfill
\begin{minipage}[t]{.66\linewidth}
\centering
\textbf{D. Larger-scale evaluation outcomes}\par\vspace{1pt}
\fontsize{8}{9}\selectfont
\setlength{\tabcolsep}{2.4pt}
\begin{tabular*}{\linewidth}{@{\extracolsep{\fill}}lcrrrr@{}}
\toprule
& & \multicolumn{2}{c}{Passed / tested} & \multicolumn{2}{c}{Best gain (\%)} \\
\cmidrule(lr){3-4}\cmidrule(lr){5-6}
Domain & S top-$k$ & S & PE & S & PE \\
\midrule
Ranking & 12 & 11/60 & \textbf{18/60} & 4.70 & \textbf{6.19} \\
Watch-time & 2 & 4/10 & \textbf{5/7} & 1.77 & \textbf{2.17} \\
RL algorithm & 3 & 0/15 & \textbf{2/12} & 1.95 & \textbf{6.64} \\
\bottomrule
\end{tabular*}
\end{minipage}
\par\vspace{1pt}
\begin{minipage}{\linewidth}\fontsize{8}{9}\selectfont
S: standalone evolution.
Bold marks the better S/PE value before rounding.
In D, S uses $k=\lceil N_{\mathrm{PE}}/5\rceil$ per source, where $N_{\mathrm{PE}}$ counts PE's larger-scale evaluations. PE evaluates promoted candidates.
Scoring and source results: Appendix~\ref{app:rl} and Table~\ref{tab:pe_source_detail}.
\end{minipage}
\end{table}

\textbf{Recommendation designs transfer to new datasets.}\quad
Under shared target-side calibration, every evaluated PE recommendation design improves on the reference in mean performance (Appendices~\ref{app:transfer-watchtime} and~\ref{app:transfer-ranking}). PE also achieves better candidate-set means on both targets (Table~\ref{tab:transfer-generalization}A).
The best individual design comes from S on VK-LSVD and PE on KuaiVideo.

For RL, the best PE and S candidates achieve 23.10\% and 22.94\% pass@1 on held-out LiveCodeBench-632 (LCB-632; Table~\ref{tab:transfer-generalization}B), both below the unmodified reference mean of 23.26\% (Appendix~\ref{app:transfer-rl}).

\begin{table}[!ht]
\centering\fontsize{8}{9}\selectfont
\setlength{\tabcolsep}{4pt}
\renewcommand{\arraystretch}{1.0}
\caption{\textbf{Transfer performance of the designs returned by each discovery procedure.}}
\label{tab:transfer-generalization}
\begin{minipage}[t]{.67\linewidth}
\centering
\textbf{A. Recommendation designs}\par\vspace{1pt}
\fontsize{8}{9}\selectfont
\setlength{\tabcolsep}{2.3pt}
\begin{tabular*}{\linewidth}{@{\extracolsep{\fill}}lrrrr@{}}
\toprule
\multirow{2}{*}{Target metric} & \multicolumn{2}{c}{Candidate-set mean} & \multicolumn{2}{c}{Best candidate} \\
\cmidrule(lr){2-3}\cmidrule(lr){4-5}
& S & PE & S & PE \\
\midrule
KuaiVideo AUC (\%) $\uparrow$ & 74.39 & \textbf{74.45} & 74.43 & \textbf{74.46} \\
VK-LSVD MAE (s) $\downarrow$ & 14.55 & \textbf{14.54} & \textbf{14.46} & 14.53 \\
\bottomrule
\end{tabular*}
\end{minipage}\hfill
\begin{minipage}[t]{.31\linewidth}
\centering
\textbf{B. RL: best candidate}\par\vspace{1pt}
\fontsize{8}{9}\selectfont
\setlength{\tabcolsep}{2pt}
\begin{tabular*}{\linewidth}{@{\extracolsep{\fill}}lrr@{}}
\toprule
LCB-632 & S & PE \\
\midrule
Pass@1 (\%) $\uparrow$ & 22.94 & \textbf{23.10} \\
\bottomrule
\end{tabular*}
\end{minipage}
\par\vspace{1pt}
\begin{minipage}{\linewidth}\fontsize{8}{9}\selectfont
Candidate scores average results across three training seeds (recommendation) or three stochastic reruns (RL). Candidate-set means average these scores across each method's evaluated recommendation designs.
Bold marks the better S/PE value before rounding; Appendix~\ref{app:postdiscovery-transfer} specifies selection and evaluation protocols.
\end{minipage}
\end{table}

\subsection{Decomposed discovery improves full-target quality and discovery yield}
\label{sec:full-results}
\label{sec:pe_track_b_decomposed_discovery_for_expensive_targets}
\label{sec:pe_compute_allocation_in_decomposed_discovery}

We compare search depth, full-target quality, and yield in completed runs under matched total GPU compute.

\textbf{Subset training sustains deeper search.}\quad
Under these budgets, PE completes 100 iterations per source search, compared with 14 LLM or 20 VLM iterations per full-target search for Direct.
Source and intermediate feedback guide candidate refinement before full-target training.

\textbf{LLM: full-target quality and completed-run yield.}\quad
PE's best full-target relative accuracy gain is 2.48\%, versus 0.92\% for Direct (Figure~\ref{fig:budget-discovery}a).
At the predefined 1\% threshold, PE yields 13 successful candidates from 39 full-target attempts; Direct yields none from 70 (Table~\ref{tab:pe_full}).
PE returns more successful candidates at all four examined thresholds from 0.5\% to 2\% (Appendix~\ref{app:budget-results}).

\textbf{VLM: gains in both model quality and discovery yield.}\quad
PE improves best full-target loss reduction from 5.05\% to 8.78\%.
PE yields 57 successful candidates from 61 full-target attempts, versus 12 from 100 for Direct (Table~\ref{tab:pe_full}).
These results support affordable subset search with screening before full-target training (final-stage outcomes across families: Appendix~\ref{app:outcomes-overview}).

\begin{table}[!ht]
\centering\fontsize{8}{9}\selectfont
\setlength{\tabcolsep}{4pt}
\renewcommand{\arraystretch}{1.0}
\caption{\textbf{Hierarchical discovery improves full-target quality and completed-run yield.}
Completed-run results use matched total GPU compute. Gains are relative LLM accuracy increases or VLM loss reductions; successful candidates meet full-target thresholds.}
\label{tab:pe_full}
\label{tab:pe_cost}
\begin{tabular*}{\linewidth}{@{\extracolsep{\fill}}llrrr@{}}
\toprule
Domain & Method & Full-target attempts & Successful candidates & Best gain (\%) \\
\midrule
\multirow{2}{*}{LLM} & Direct & 70 & 0 & 0.92 \\
& PE & 39 & \textbf{13} & \textbf{2.48} \\
\midrule
\multirow{2}{*}{VLM} & Direct & 100 & 12 & 5.05 \\
& PE & 61 & \textbf{57} & \textbf{8.78} \\
\bottomrule
\end{tabular*}
\par\vspace{1pt}
\begin{minipage}{\linewidth}\fontsize{8}{9}\selectfont
Figure~\ref{fig:budget-discovery} summarizes LLM quality and yield from these completed runs.
Appendices~\ref{app:execution-accounting} and~\ref{app:vlm-checkpoint} detail budget accounting, evaluation counts, and mean scores of passing evaluations.
\end{minipage}
\end{table}

\textbf{Successful LLM designs span several intervention types.}\quad
\label{sec:idea-diversity}
The 13 successful LLM candidates span changes to Muon update geometry, RoPE configurations within the existing hybrid attention stack, and late-training checkpoint averaging.
Section~\ref{sec:idea-cases} follows one discovery; Appendix~\ref{app:mechanism-analysis} groups the mechanisms and lists designs from all five families (Table~\ref{tab:graduated-designs}).

\subsection{Applicability across search harnesses and proposal models}
\label{sec:pe_applicability_across_harnesses_and_proposal_models}

For watch-time discovery from EGMN, we vary the harness or proposer while holding the other fixed in matched PE/S comparisons (Appendix~\ref{app:harness-proposers}, Table~\ref{tab:pe_compat}).
PE improves the best larger-scale gain with 4/5 harnesses and, with AdaEvolve fixed, all four proposers. With Claude Opus 5, the gain rises from 2.61\% to 3.64\%.
PE improves mean best local gains with 4/5 harnesses and 2/4 proposers. With both Opus proposers, PE has lower local means but better larger-scale outcomes than S; it improves best local gains on at least three of five datasets with every proposer.

\section{Analysis: How Collaboration Supports Discovery}
\label{sec:analysis}

We compare local S--PE proposals with and without shared feedback (Section~\ref{sec:joint-results}), measure direct reuse through peer replay (Section~\ref{sec:transfer-analysis}), and examine reconstructed traces of evidence availability and subsequent refinement (Section~\ref{sec:idea-cases}).

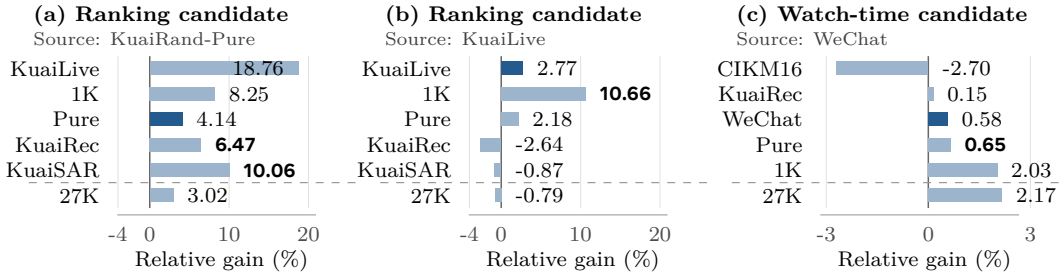
\begin{figure}[!ht]
\centering
\begin{tikzpicture}[x=1cm,y=.70cm,font=\fontsize{8}{9.5}\selectfont]
\begin{scope}[shift={(0.0,0)}]
\node[anchor=west,font=\footnotesize\bfseries] at (.04,.99) {(a) Ranking candidate};
\node[anchor=west,font=\scriptsize,text=black!75] at (.04,.57) {Source: KuaiRand-Pure};
\draw[black!12,line width=.35pt] (1.2800,.23)--(1.2800,-2.64);
\node[anchor=north,text=black!70] at (1.2800,-2.82) {-4};
\draw[black!12,line width=.35pt] (1.7000,.23)--(1.7000,-2.64);
\node[anchor=north,text=black!70] at (1.7000,-2.82) {0};
\draw[black!12,line width=.35pt] (2.7500,.23)--(2.7500,-2.64);
\node[anchor=north,text=black!70] at (2.7500,-2.82) {10};
\draw[black!12,line width=.35pt] (3.8000,.23)--(3.8000,-2.64);
\node[anchor=north,text=black!70] at (3.8000,-2.82) {20};
\draw[black!60,line width=.55pt] (1.7000,.23)--(1.7000,-2.64);
\node[anchor=east] at (1.16,0.0000) {KuaiLive};
\fill[peblue!45] (1.7000,-0.1300) rectangle (3.6698,0.1300);
\node[anchor=east] at (3.60,0.0000) {18.76};
\node[anchor=east] at (1.16,-0.4800) {1K};
\fill[peblue!45] (1.7000,-0.6100) rectangle (2.5663,-0.3500);
\node[anchor=west] at (2.6263,-0.4800) {8.25};
\node[anchor=east] at (1.16,-0.9600) {Pure};
\fill[peblue] (1.7000,-1.0900) rectangle (2.1347,-0.8300);
\node[anchor=west] at (2.1947,-0.9600) {4.14};
\node[anchor=east] at (1.16,-1.4400) {KuaiRec};
\fill[peblue!45] (1.7000,-1.5700) rectangle (2.3793,-1.3100);
\node[anchor=west] at (2.4393,-1.4400) {\textbf{6.47}};
\node[anchor=east] at (1.16,-1.9200) {KuaiSAR};
\fill[peblue!45] (1.7000,-2.0500) rectangle (2.7563,-1.7900);
\node[anchor=west] at (2.8163,-1.9200) {\textbf{10.06}};
\node[anchor=east] at (1.16,-2.4000) {27K};
\fill[peblue!45] (1.7000,-2.5300) rectangle (2.0171,-2.2700);
\node[anchor=west] at (2.0771,-2.4000) {3.02};
\draw[black!40,dashed,line width=.45pt] (.10,-2.16)--(4.45,-2.16);
\draw[black!40,line width=.4pt] (1.27,-2.77)--(3.90,-2.77);
\node[anchor=north] at (2.65,-3.25) {Relative gain (\%)};
\end{scope}
\begin{scope}[shift={(4.65,0)}]
\node[anchor=west,font=\footnotesize\bfseries] at (.04,.99) {(b) Ranking candidate};
\node[anchor=west,font=\scriptsize,text=black!75] at (.04,.57) {Source: KuaiLive};
\draw[black!12,line width=.35pt] (1.2800,.23)--(1.2800,-2.64);
\node[anchor=north,text=black!70] at (1.2800,-2.82) {-4};
\draw[black!12,line width=.35pt] (1.7000,.23)--(1.7000,-2.64);
\node[anchor=north,text=black!70] at (1.7000,-2.82) {0};
\draw[black!12,line width=.35pt] (2.7500,.23)--(2.7500,-2.64);
\node[anchor=north,text=black!70] at (2.7500,-2.82) {10};
\draw[black!12,line width=.35pt] (3.8000,.23)--(3.8000,-2.64);
\node[anchor=north,text=black!70] at (3.8000,-2.82) {20};
\draw[black!60,line width=.55pt] (1.7000,.23)--(1.7000,-2.64);
\node[anchor=east] at (1.16,0.0000) {KuaiLive};
\fill[peblue] (1.7000,-0.1300) rectangle (1.9909,0.1300);
\node[anchor=west] at (2.0509,0.0000) {2.77};
\node[anchor=east] at (1.16,-0.4800) {1K};
\fill[peblue!45] (1.7000,-0.6100) rectangle (2.8193,-0.3500);
\node[anchor=west] at (2.8793,-0.4800) {\textbf{10.66}};
\node[anchor=east] at (1.16,-0.9600) {Pure};
\fill[peblue!45] (1.7000,-1.0900) rectangle (1.9289,-0.8300);
\node[anchor=west] at (1.9889,-0.9600) {2.18};
\node[anchor=east] at (1.16,-1.4400) {KuaiRec};
\fill[peblue!45] (1.7000,-1.5700) rectangle (1.4228,-1.3100);
\node[anchor=west] at (1.7700,-1.4400) {-2.64};
\node[anchor=east] at (1.16,-1.9200) {KuaiSAR};
\fill[peblue!45] (1.7000,-2.0500) rectangle (1.6086,-1.7900);
\node[anchor=west] at (1.7700,-1.9200) {-0.87};
\node[anchor=east] at (1.16,-2.4000) {27K};
\fill[peblue!45] (1.7000,-2.5300) rectangle (1.6170,-2.2700);
\node[anchor=west] at (1.7700,-2.4000) {-0.79};
\draw[black!40,dashed,line width=.45pt] (.10,-2.16)--(4.45,-2.16);
\draw[black!40,line width=.4pt] (1.27,-2.77)--(3.90,-2.77);
\node[anchor=north] at (2.65,-3.25) {Relative gain (\%)};
\end{scope}
\begin{scope}[shift={(9.3,0)}]
\node[anchor=west,font=\footnotesize\bfseries] at (.04,.99) {(c) Watch-time candidate};
\node[anchor=west,font=\scriptsize,text=black!75] at (.04,.57) {Source: WeChat};
\draw[black!12,line width=.35pt] (1.3500,.23)--(1.3500,-2.64);
\node[anchor=north,text=black!70] at (1.3500,-2.82) {-3};
\draw[black!12,line width=.35pt] (2.7000,.23)--(2.7000,-2.64);
\node[anchor=north,text=black!70] at (2.7000,-2.82) {0};
\draw[black!12,line width=.35pt] (4.0500,.23)--(4.0500,-2.64);
\node[anchor=north,text=black!70] at (4.0500,-2.82) {3};
\draw[black!60,line width=.55pt] (2.7000,.23)--(2.7000,-2.64);
\node[anchor=east] at (1.16,0.0000) {CIKM16};
\fill[peblue!45] (2.7000,-0.1300) rectangle (1.4850,0.1300);
\node[anchor=west] at (2.7700,0.0000) {-2.70};
\node[anchor=east] at (1.16,-0.4800) {KuaiRec};
\fill[peblue!45] (2.7000,-0.6100) rectangle (2.7675,-0.3500);
\node[anchor=west] at (2.8275,-0.4800) {0.15};
\node[anchor=east] at (1.16,-0.9600) {WeChat};
\fill[peblue] (2.7000,-1.0900) rectangle (2.9610,-0.8300);
\node[anchor=west] at (3.0210,-0.9600) {0.58};
\node[anchor=east] at (1.16,-1.4400) {Pure};
\fill[peblue!45] (2.7000,-1.5700) rectangle (2.9925,-1.3100);
\node[anchor=west] at (3.0525,-1.4400) {\textbf{0.65}};
\node[anchor=east] at (1.16,-1.9200) {1K};
\fill[peblue!45] (2.7000,-2.0500) rectangle (3.6135,-1.7900);
\node[anchor=west] at (3.6735,-1.9200) {2.03};
\node[anchor=east] at (1.16,-2.4000) {27K};
\fill[peblue!45] (2.7000,-2.5300) rectangle (3.6765,-2.2700);
\node[anchor=west] at (3.7365,-2.4000) {2.17};
\draw[black!40,dashed,line width=.45pt] (.10,-2.16)--(4.45,-2.16);
\draw[black!40,line width=.4pt] (1.27,-2.77)--(3.90,-2.77);
\node[anchor=north] at (2.65,-3.25) {Relative gain (\%)};
\end{scope}
\end{tikzpicture}
\caption{\textbf{Peer discoveries can outperform local candidates; larger-scale evaluation tests transfer.}
Each panel shows one design retrained unchanged across six instances.
Gains use fixed instance references; bold values exceed the best gain from locally proposed PE candidates on the receiving instance (Table~\ref{tab:pe_joint}).
Dark bars identify sources; dashed lines separate larger-scale evaluations.
1K/Pure/27K abbreviate KuaiRand datasets (Appendix~\ref{app:mechanism-analysis}).}
\label{fig:transfer-profiles}
\end{figure}

\subsection{Peer discoveries support direct reuse and inform promotion}
\label{sec:transfer-analysis}

\textbf{Direct reuse improves results beyond locally proposed candidates.}\quad
We compare PE's best locally proposed candidate (Table~\ref{tab:pe_joint}A--B) with its best available candidate after replaying peer discoveries on the same instance.
Including replayed peer candidates improves PE's best available gains on three of five ranking and one of five watch-time datasets, raising PE's mean best gains from 8.97\% to 11.29\% and 3.85\% to 3.95\%, respectively (Appendix~\ref{app:source-outcomes}, Table~\ref{tab:pe_source_detail}).
A KuaiRand-Pure ranking candidate raises KuaiSAR's best available PE gain from 3.88\% to 10.06\%; a WeChat watch-time candidate raises KuaiRand-Pure's from 0.16\% to 0.65\%.
Replay results inform proposals, while each search continues to select parents from its local archive (Section~\ref{sec:ideas}).
Benefits vary by dataset; with peer candidates, the mean PE--S difference stays positive when any dataset is omitted (Appendix~\ref{app:source-outcomes}).

\textbf{Cross-instance and larger-scale evidence guide which designs advance.}\quad
Figure~\ref{fig:transfer-profiles} compares three individual PE designs, each retrained unchanged on its source, four peers, and larger-scale KuaiRand-27K.
The KuaiRand-Pure ranking candidate improves all four peers and gains 3.02\% at larger scale (a).
The KuaiLive ranking candidate improves KuaiRand-1K by 10.66\% but declines by 0.79\% at larger scale (b).
The WeChat watch-time head reduces larger-scale MAE by 2.17\% despite regressing on CIKM16 (c).
All three pass peer screening, but only two retain positive gains at larger scale.
Peer evaluation identifies reusable designs; larger-scale evaluation checks whether their gains persist.

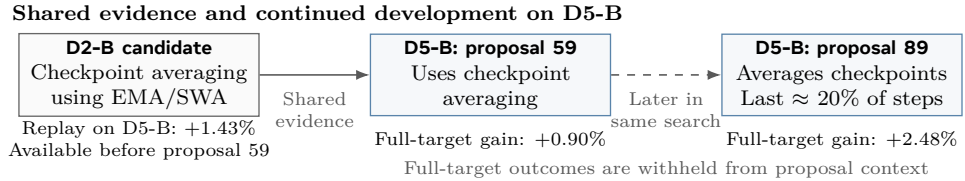
\begin{figure}[!ht]
\centering
\begin{tikzpicture}[x=1cm,y=1cm,font=\footnotesize]
\node[anchor=west,font=\footnotesize\bfseries] at (.05,4.6) {Shared evidence and continued development on D5-B};
\node[draw=black!60,fill=black!2,line width=.5pt,align=center,text width=3.0cm,minimum height=1.05cm,inner sep=2.5pt] (origin) at (1.85,3.80) {\textbf{D2-B candidate}\\Checkpoint averaging\\using EMA/SWA};
\node[draw=peblue!75,fill=peblue!4,line width=.5pt,align=center,text width=3.0cm,minimum height=1.05cm,inner sep=2.5pt] (adopt) at (6.50,3.80) {\textbf{D5-B: proposal 59}\\Uses checkpoint\\averaging};
\node[draw=peblue!75,fill=peblue!4,line width=.5pt,align=center,text width=3.0cm,minimum height=1.05cm,inner sep=2.5pt] (refine) at (11.15,3.80) {\textbf{D5-B: proposal 89}\\Averages checkpoints\\Last $\approx20\%$ of steps};
\draw[-{Latex[length=1.8mm]},black!65,line width=.6pt] (origin.east)--node[below=3pt,align=center,font=\scriptsize]{Shared\\evidence}(adopt.west);
\draw[-{Latex[length=1.8mm]},black!65,dashed,line width=.6pt] (adopt.east)--node[below=3pt,align=center,font=\scriptsize]{Later in\\same search}(refine.west);
\node[align=center,font=\scriptsize] at (1.85,2.94) {Replay on D5-B: $+1.43\%$\\Available before proposal 59};
\node[align=center,font=\scriptsize] at (6.50,2.94) {Full-target gain: $+0.90\%$};
\node[align=center,font=\scriptsize] at (11.15,2.94) {Full-target gain: $+2.48\%$};
\node[align=center,font=\scriptsize,text=black!65] at (8.83,2.55) {Full-target outcomes are withheld from proposal context};
\end{tikzpicture}
\caption{\textbf{Shared checkpoint-averaging evidence and subsequent local development.}
D2-B evidence is available before D5-B proposal 59; proposal 89 later refines the averaging window.
Arrows indicate reconstructed evidence flow and temporal order, not direct parentage. Gains are retrospective comparisons with fixed instance references.
Transfer results: Figure~\ref{fig:checkpoint-transfer} (Appendix~\ref{app:case-llm}).}
\label{fig:idea-cases}
\end{figure}

\subsection{Tracing shared evidence and subsequent local refinement}
\label{sec:idea-cases}

The LLM case reconstructs available shared evidence and traces subsequent local adaptations of established techniques; a complementary watch-time reuse case appears in Appendix~\ref{app:watch-reuse-case}.

\textbf{Shared checkpoint averaging and continued local development.}\quad

In LLM pretraining, a D2-B candidate implements checkpoint averaging using exponential moving average (EMA) and stochastic weight averaging (SWA)~\citep{izmailov2018swa}.
Replay yields a 1.43\% relative accuracy gain on D5-B. Reconstructing context from archived records and the selection rule indicates availability before proposal 59.
Proposal 59 uses checkpoint averaging late in training.
Proposal 89 later restricts averaging to approximately the final 20\% of training.
Proposals 59 and 89 gain 0.90\% and 2.48\%, respectively, over the fixed full-target reference. Proposal 89 raises accuracy from its direct parent's 41.13\% to 41.25\% (Table~\ref{tab:case-parent-comparisons}).
Full-target outcomes were withheld from proposal feedback.

\section{Related Work}
\label{sec:related-work}

Architecture search, Population-Based Training, and AutoML-Zero automate model and algorithm design~\citep{real2019regularized,jaderberg2017pbt,real2020automlzero}.
LLM-guided search enables flexible code edits~\citep{romeraparedes2024funsearch,liu2024eoh,novikov2025alphaevolve,lange2025shinkaevolve}, while quality-diversity methods maintain diverse candidates~\citep{mouret2015mapelites,nasir2024llmatic}.
Evolutionary multitasking and multi-task architecture search transfer knowledge across tasks~\citep{gupta2016multifactorial,liao2024moemtnas,zhang2025ktnas}; AgentRxiv shares research reports~\citep{schmidgall2025agentrxiv}.
FunSearch uses islands to maintain diverse populations for one target problem~\citep{romeraparedes2024funsearch}. EMO-STA evolves a shared archive before adapting selected programs to individual tasks~\citep{gozeten2026emosta}.
PE maintains ongoing task-specific searches and archives, sharing executable interventions and peer evaluation results to guide local edits and cross-scale promotion.

NASNet transfers architectures from CIFAR-10 to ImageNet~\citep{zoph2018nasnet}, while FABOLAS varies dataset size during hyperparameter optimization~\citep{klein2017fabolas}.
AlphaEvolve uses evaluation cascades~\citep{novikov2025alphaevolve}; multi-fidelity methods allocate evaluation resources~\citep{li2018hyperband,trofimov2020multifidelity}, and Janus co-evolves proxy evaluators~\citep{liu2026janus}.
PE combines affordable local search, peer and intermediate screening, and full-target verification.

Existing benchmarks cover competitions and ML experimentation~\citep{chan2024mlebench,nathani2025mlgym}, multi-GPU/multi-node experiments~\citep{wu2025innovatorbench}, and paper-derived research tasks in ResearchGym~\citep{garikaparthi2026researchgym}.
Others cover post-training~\citep{rank2026posttrainbench}, adaptive pretraining budgets~\citep{lyu2026mlsbench}, or open-ended tasks~\citep{zhu2026edgebench}.
RMD-Bench evaluates collaboration across ongoing local searches and discovery across training scales, including transfer to new datasets and tasks in Track~A and full-target quality and yield in Track~B (Table~\ref{tab:benchmark_comparison}).

\section{Conclusion}
\label{sec:conclusion}
Population Evolution connects searches through shared evidence and hierarchical promotion; RMD-Bench supplies executable evaluation protocols.
Shared feedback improves mean best local recommendation gains; peer replay enables reuse. PE improves best larger-scale outcomes, and its recommendation designs transfer to new datasets.
Under matched total GPU compute, hierarchical discovery improves full-target LLM/VLM quality and completed-run yield.
Watch-time best larger-scale gains improve with four of five harnesses and all four proposers.
Appendix~\ref{app:limitations} discusses limitations and future directions.

\clearpage
\subsection*{AI use statement}
The authors wrote most of the manuscript and supplied its key scientific points, numerical results, and technical details.
Generative AI tools were used in a supporting role to draft and polish selected passages, retrieve related literature, and provide feedback on methodology and result interpretation.
LLM-driven proposal and code generation within the discovery experiments are described in Section~\ref{sec:method} and Appendix~\ref{app:implementation}.
The authors reviewed and revised AI-assisted content and made the final scientific and editorial decisions, taking full responsibility for the final text, claims, results, and research artifacts.

\subsection*{Reproducibility statement}
Appendix~\ref{app:datasets} specifies the datasets, preprocessing, starting models, editable components, training and evaluation protocols, and baseline calibration used in RMD-Bench.
Appendix~\ref{app:implementation} documents PE's publication and promotion criteria, shared proposal context, prompts, asynchronous search procedure, and evaluation accounting.
Appendix~\ref{app:postdiscovery-transfer} details candidate selection, target calibration, reference scores, and repeated transfer evaluations.
Appendices~\ref{app:additional-details} and~\ref{app:cross-implementations} report per-instance outcomes and additional configuration and budget comparisons.
The benchmark and search code will be made publicly available.

\bibliography{references}
\bibliographystyle{iclr2027_conference}

\clearpage
\appendix
\section{RMD-Bench Specifications and Evaluation Protocols}
\label{app:datasets}

This appendix specifies the data, starting model, editable interface, and evaluation protocol for each task family in Section~\ref{sec:benchmark}.
Table~\ref{tab:pe_bench} summarizes the five families; the following subsections give their protocols, baseline calibration, and discovery compute scales.
Counts describe the processed inputs used in our experiments and can differ from the original public releases.
Following Section~\ref{sec:method}, source instances host local search, intermediate instances provide Track~B screening and feedback, and larger-scale or full-target instances assess selected designs.
Each PE source search receives 100 iterations.
Appendix~\ref{app:postdiscovery-transfer} specifies the additional evaluations of selected designs beyond these discovery instances.
For error and loss objectives, a harness that maximizes fitness receives the negative metric; reported values and promotion retain the original lower-is-better metric.

\begin{table}[H]
\centering
\footnotesize
\setlength{\tabcolsep}{3pt}
\renewcommand{\arraystretch}{1.17}
\caption{\textbf{RMD-Bench task families and discovery spaces.}
Each family provides five source instances.
Track~A adds a larger-scale evaluation instance; Track~B adds two intermediate instances and a full target.
Sizes refer to the reference implementations and processed corpora used here.
Detailed task specifications follow.}
\label{tab:pe_bench}
\begin{tabular}{@{}P{1.4cm}P{2.6cm}P{2.8cm}P{3.0cm}P{2.95cm}@{}}
\toprule
\textbf{Task family} & \textbf{Starting method} & \textbf{Discovery space} & \textbf{Source instances} & \textbf{Final-stage evaluation} \\
\midrule
\multicolumn{5}{@{}l}{\textit{Track A: joint discovery across related instances}} \\
\addlinespace[2pt]
Ranking & FuXi-Linear~\citep{ye2026fuxilinear} & Sequence representations; temporal and retention mechanisms & KuaiLive, KuaiRand-1K/Pure, KuaiRec, KuaiSAR & KuaiRand-27K; 146.82M interactions; ranking score $\uparrow$ \\
\addlinespace[3pt]
Watch-time & EGMN~\citep{zhao2025egmn} & Distributional head; duration prediction & CIKM16, KuaiRec, WeChat, KuaiRand-Pure/1K & KuaiRand-27K; 23.99M examples; MAE (s) $\downarrow$ \\
\addlinespace[3pt]
RL algorithm & Qwen2.5-Coder~\citep{hui2024qwen25coder}; GRPO & Policy loss; rollout filtering & Five sources, different seeds; 1.5B policy, 15 steps & 3B policy, 50 steps; 423 LiveCodeBench problems; pass@1 $\uparrow$ \\
\midrule
\multicolumn{5}{@{}l}{\textit{Track B: decomposed discovery for expensive targets}} \\
\addlinespace[2pt]
LLM pretraining & Gemma~3-style~\citep{gemmateam2025gemma3}; $\sim$185M & Architecture; optimizer; schedule; training mechanisms & FineWeb-Edu 100B core set; two 2\% and three 5\% views & Full core set; seven-task accuracy $\uparrow$ \\
\addlinespace[3pt]
VLM pretraining & Qwen3.5-style~\citep{qwenteam2026qwen35} + SigLIP; $\sim$646M trainable & Language architecture; projector; optimizer; schedule & FineVisionMax format; two 2\% and three 5\% views & 2.38M raw training examples; validation loss $\downarrow$ \\
\bottomrule
\end{tabular}
\end{table}

\FloatBarrier

\subsection{Sequential ranking}
\label{app:ranking}

\paragraph{Data and preprocessing.} The source family comprises KuaiLive~\citep{qu2025kuailive}, KuaiRand-1K and KuaiRand-Pure~\citep{gao2022kuairand}, KuaiRec~\citep{gao2022kuairec}, and KuaiSAR~\citep{sun2023kuaisar}.
KuaiLive treats clicked streamers as items.
KuaiSAR uses clicked interactions from recommendation pages.
For each user or session, interactions are ordered chronologically; the last interaction is the next-item target, and preceding interactions form the history.
KuaiRand ranking uses the processed standard-exposure logs.
The benchmark evaluates full-catalog next-item ranking.

\begin{table}[htbp]
\centering\footnotesize
\setlength{\tabcolsep}{3pt}
\caption{Processed ranking datasets and training allocation.
All but KuaiRand-27K are source instances.
Batch size is the per-GPU value selected by baseline HPO (Table~\ref{tab:ranking_hpo}).}
\label{tab:ranking_data}
\begin{tabular}{@{}lrrrrrrr@{}}
\toprule
Dataset & Users & Interactions & Items & \shortstack{Mean seq.\\length} & \shortstack{Context\\length} & \shortstack{Batch size\\per GPU} & Epochs \\
\midrule
KuaiLive & 22,497 & 4,381,721 & 151,849 & 194.77 & 512 & 47 & 30 \\
KuaiRand-1K & 988 & 290,969 & 2,025 & 294.50 & 1,024 & 38 & 30 \\
KuaiRand-Pure & 25,220 & 1,235,423 & 6,073 & 48.99 & 512 & 88 & 30 \\
KuaiRec & 7,176 & 12,529,113 & 9,958 & 1,745.97 & 1,024 & 40 & 50 \\
KuaiSAR & 17,491 & 1,595,760 & 55,505 & 91.23 & 512 & 46 & 50 \\
\midrule
KuaiRand-27K & 27,285 & 146,821,616 & 461,994 & 5,381.04 & 1,024 & 51 & 50 \\
\bottomrule
\end{tabular}
\end{table}

\paragraph{Starting model and training.} The reference is FuXi-Linear~\citep{ye2026fuxilinear}, with positional and temporal channels and a causal chunkwise implementation (Table~\ref{tab:ranking_reference_config}).
User and item embeddings are L2-normalized and scored by dot product.
Training uses autoregressive sampled softmax and AdamW.
Baseline HPO selects dropout, learning rate, per-GPU batch size, and weight decay separately for each dataset (Appendix~\ref{app:baseline-construction}).
Table~\ref{tab:ranking_data} gives the resulting training allocations, and Table~\ref{tab:ranking_hpo} lists the selected hyperparameters.

\begin{table}[htbp]
\centering\small
\setlength{\tabcolsep}{5pt}
\caption{Fixed FuXi-Linear reference configuration.}
\label{tab:ranking_reference_config}
\begin{tabular}{@{}P{4.2cm}P{8.3cm}@{}}
\toprule
Component & Setting \\
\midrule
Encoder & Four blocks; four attention heads; $d_q=d_k=d_v=32$ \\
Embeddings and position & Item dimension 128; positional linear channel dimension 32 \\
Temporal channel & Eight heads; base 2, stride 3, start index 10; learnable decay \\
Sequence computation & Chunk size 128; RoPE disabled \\
Sampled-softmax loss & 128 negatives; temperature 0.05 \\
Execution & Eight-GPU DDP; BF16/TF32 \\
\bottomrule
\end{tabular}
\end{table}

\begin{samepage}
\paragraph{Editable interface.}
The proposer can jointly modify three components:
\begin{itemize}\raggedright
\item FuXi-Linear blocks, channels, and the user representation.
\item Retention, temporal interactions, normalization, and gating.
\item The causal chunkwise attention kernel.
\end{itemize}
\end{samepage}
The input/output interfaces and the training path from user representations to sampled-softmax loss remain intact.
Data loading, loss, sampling, and evaluation are fixed.
KuaiRand-1K additionally uses the per-candidate HPO procedure in Appendix~\ref{app:baseline-construction}; the other four sources use fixed optimizer settings.

\paragraph{Evaluation and promotion.} The ranking score used as evolutionary fitness is
\[
R_{\mathrm{rank}}=(\mathrm{NDCG@10}+\mathrm{HR@10}+\mathrm{MRR})/3.
\]
Published interventions are retrained on the four peer datasets, and candidates meeting the peer criterion advance to KuaiRand-27K for larger-scale evaluation.
Table~\ref{tab:pe_gates} specifies publication, peer, and larger-scale success thresholds; standalone top-$k$ candidates use the same larger-scale evaluator and success criterion.

\subsection{Watch-time prediction}
\label{app:watch}

\paragraph{Data and preprocessing.} The source datasets are CIKM16~\citep{cikm2016challenge}, KuaiRec~\citep{gao2022kuairec}, WeChat, KuaiRand-Pure, and KuaiRand-1K~\citep{gao2022kuairand}, with KuaiRand-27K as the larger-scale evaluation instance.
Following EGMN~\citep{zhao2025egmn}, we include CIKM16 as a related engagement-duration task: its labels describe time spent in online search sessions.
Training uses a fixed deterministic 80\% partition; the remaining 20\% supplies the evaluation scores used throughout discovery.
Within each dataset, all methods share a p99 label scale estimated from the processed corpus before partitioning.
KuaiRand watch-time preprocessing combines random- and standard-exposure logs.
KuaiRec and KuaiRand-27K use sampled processed corpora.
Table~\ref{tab:watch_data} reports processed examples rather than original-release interactions.

\begin{table}[htbp]
\centering\footnotesize
\setlength{\tabcolsep}{3pt}
\caption{Processed watch-time datasets.
The user/query vocabulary counts queries for CIKM16 and users otherwise; the p99 scale converts normalized MAE to seconds.
These counts reflect watch-time preprocessing and differ from ranking.}
\label{tab:watch_data}
\begin{tabular}{@{}lrrrrrr@{}}
\toprule
Dataset & Train & Eval. & Total & \shortstack{User/query\\vocabulary} & \shortstack{Item\\vocabulary} & \shortstack{p99 scale\\(s)} \\
\midrule
CIKM16 & 246,853 & 61,714 & 308,567 & 9,837 & 122,642 & 5.237 \\
KuaiRec & 980,676 & 245,169 & 1,225,845 & 7,176 & 9,599 & 47.697 \\
WeChat & 5,059,605 & 1,264,902 & 6,324,507 & 19,994 & 95,511 & 165.364 \\
KuaiRand-Pure & 1,871,608 & 467,902 & 2,339,510 & 27,279 & 7,268 & 182.240 \\
KuaiRand-1K & 6,951,281 & 1,737,821 & 8,689,102 & 1,000 & 2,926,199 & 188.515 \\
\midrule
KuaiRand-27K & 19,193,265 & 4,798,317 & 23,991,582 & 27,285 & 5,489,023 & 187.368 \\
\bottomrule
\end{tabular}
\end{table}

\paragraph{Starting model and training.} EGMN~\citep{zhao2025egmn} uses 16-dimensional sparse and sequence embeddings, mean-pooled sequence features, linear projections of continuous features, and a 256--128--64 MLP encoder.
Its starting head combines one exponential component for fast skips and ten non-negative left-truncated Gaussian components.
The point prediction is the mixture expectation.
The starting model's objective combines the mixture negative log-likelihood with entropy and absolute-error terms weighted by $\alpha$ and $\beta$:
\[
\mathcal L=\mathcal L_{\mathrm{mixture\ NLL}}+\alpha\mathcal L_{\mathrm{entropy}}+\beta|\hat y-y|.
\]
Runs use Adagrad, 10 epochs, batch size 2,048, and a fixed training seed.
The S and PE runs with GPT-5.5 share the HPO-selected configurations in Table~\ref{tab:watch_hpo}, with per-candidate HPO disabled; Appendix~\ref{app:baseline-construction} specifies calibration.

\begin{table}[htbp]
\centering\small
\setlength{\tabcolsep}{4pt}
\caption{Fixed HPO-selected watch-time hyperparameters shared by the S and PE runs with GPT-5.5.}
\label{tab:watch_hpo}
\begin{tabular}{@{}lrrrrr@{}}
\toprule
Dataset & Learning rate & Weight decay & $\alpha$ & $\beta$ & Dropout \\
\midrule
CIKM16 & 0.091684 & $3.35\times10^{-6}$ & 0.659274 & 2.023865 & 0.141097 \\
KuaiRec & 0.25076 & $1.94\times10^{-5}$ & 0.619568 & 3.41108 & 0.059406 \\
WeChat & 0.085141 & $1.98\times10^{-6}$ & 0.454659 & 3.263077 & 0.055439 \\
KuaiRand-Pure & 0.064052 & $8.58\times10^{-7}$ & 0.521568 & 4.296266 & 0.077754 \\
KuaiRand-1K & 0.134232 & $2.31\times10^{-5}$ & 0.646871 & 1.511465 & 0.362471 \\
KuaiRand-27K & 0.102368 & $1.53\times10^{-5}$ & 0.489522 & 1.778713 & 0.138667 \\
\bottomrule
\end{tabular}
\end{table}

\paragraph{Editable interface and fitness.} Only the watch-time prediction head is evolvable.
Its search space includes likelihood parameterization, mixture components, point prediction, and duration conditioning.
All candidates share the feature encoder, data pipeline, optimizer settings, epoch count, batch size, and evaluator.
The evaluation metric is MAE (lower is better), reported in seconds on this partition using the fixed p99 label scale in Table~\ref{tab:watch_data}.
Candidates advance through peer and larger-scale evaluation using the MAE-reduction criteria in Table~\ref{tab:pe_gates}.

\subsection{RL algorithm discovery}
\label{app:rl}

\paragraph{Source and larger-scale instances.} The five source instances differ only in the training random seed.
Each uses Qwen2.5-Coder-1.5B-Instruct~\citep{hui2024qwen25coder}, 15 GRPO steps, and the same 408-problem DeepCoder Codeforces/test partition~\citep{luo2025deepcoder}.
The larger-scale instance uses Qwen2.5-Coder-3B-Instruct, 50 steps, and a disjoint 423-problem LiveCodeBench partition~\citep{jain2024livecodebench}.
This partition is a deterministic platform-by-difficulty stratified selection from a 1,055-problem snapshot.
Table~\ref{tab:rl_contract} summarizes the source and larger-scale training and evaluation settings.

The RL training pool has 16,252 DeepCoder \texttt{primeintellect/train} examples and 7,436 \texttt{taco/train} examples, totaling 23,688 problems.
After normalizing prompt text, we find no exact overlap between the training and evaluation prompts and retain all training examples.
Every candidate starts from the same pretrained policy weights, without inheriting trained weights from earlier search iterations.

\begin{table}[htbp]
\centering\small
\setlength{\tabcolsep}{6pt}
\caption{Fixed RL training and evaluation contracts.
Both training scales use eight responses per prompt and 16 H200 GPUs per candidate evaluation.}
\label{tab:rl_contract}
\begin{tabular}{@{}lrr@{}}
\toprule
Property & Source (each of five instances) & Larger scale \\
\midrule
Policy parameters & 1.5B & 3B \\
GRPO steps & 15 & 50 \\
Prompts per step & 64 & 32 \\
Evaluation problems & 408 & 423 \\
Evaluation tests & Public tests & Private tests \\
Training seed & Instance-specific & Fixed \\
\bottomrule
\end{tabular}

\par\vspace{3pt}
\begin{minipage}{\linewidth}\footnotesize
Source instances use tensor parallelism 2 and maximum response length 4,000.
The larger-scale partition contains 241 AtCoder, four Codeforces, and 178 LeetCode problems.
\end{minipage}
\end{table}

\paragraph{Training and search space.} The RL protocol disables HPO and fixes Adam with learning rate $10^{-6}$, a low-variance KL estimator with coefficient 0.001, and clipping coefficients 0.2/0.28.

\paragraph{Editable interface.} Evolution can modify the GRPO loss through changes to surrogate-loss geometry, clipping treatment, KL treatment, and advantage weighting, and can modify rollout filtering through the selection, weighting, or rejection of rollout groups.
It cannot change the model, training corpus, learning rate, fixed clipping/KL coefficients, RL horizon, rollout batch, response count, or evaluator.

\paragraph{Two reward levels.} Within candidate training, a sampled program receives the fraction of unit tests passed, with a $-0.1$ penalty for unparseable or truncated output.
The outer evolutionary fitness is code-generation pass@1 under deterministic decoding: the number of solved evaluation problems divided by the partition size.
Evaluation generates one completion per problem, with the evaluator's test cases excluded from the generation prompt.
The larger-scale evaluator uses hidden private tests.
The fixed reference score is the arithmetic mean of three valid baseline repetitions.

\paragraph{Promotion and reported outcomes.} Table~\ref{tab:pe_gates} gives the joint relative and absolute improvement criteria for publication, peer validation, and larger-scale success.
The larger-scale evaluation measures transfer jointly across policy size, training horizon, and code benchmark.
The five RL sources share the same task, training data, and objective and differ only in random seed. We therefore treat them as parallel searches for the strongest algorithm and report the best outcome across the five searches.

\subsection{LLM pretraining}
\label{app:llm}

\paragraph{Data hierarchy.} The benchmark uses the FineWeb-Edu 100B core set from the shuffled release~\citep{lozhkov2024finewebedu,karpathy2025finewebedu100b}.
Table~\ref{tab:llm_data} gives the exact corpus inventories and training allocations.
Source and intermediate subsets use fixed contiguous shard ranges from the shuffled corpus.
The three 5\% sources D5-A/B/C are mutually disjoint; D2-A/B are 2\% subsets nested within D5-A/B, respectively.
D10 clean is disjoint from every source slice.
D10+C combines D5-C with a fresh disjoint 5\% slice, testing an increase in data around a known source.
The full-target instance trains on the complete reference corpus.

\begin{table}[htbp]
\centering\small
\setlength{\tabcolsep}{4pt}
\caption{LLM training hierarchy.
Token counts describe the processed corpus; step counts scale approximately with the nominal corpus fractions.
All instances use the same seven-task mean accuracy.}
\label{tab:llm_data}
\begin{tabular}{@{}llrrrr@{}}
\toprule
Dataset & Role & Shards & Documents & Tokens & Steps \\
\midrule
D2-A & Source & 36 & 1,924,096 & 1,925,811,698 & 694 \\
D2-B & Source & 36 & 1,915,904 & 1,921,710,324 & 694 \\
D5-A & Source & 91 & 4,853,760 & 4,862,315,015 & 1,750 \\
D5-B & Source & 91 & 4,844,544 & 4,859,363,728 & 1,750 \\
D5-C & Source & 91 & 4,853,760 & 4,863,973,228 & 1,750 \\
D10 clean & Intermediate & 182 & 9,712,640 & 9,727,423,143 & 3,470 \\
D10+C & Intermediate & 182 & 9,711,616 & 9,728,203,396 & 3,470 \\
Full target & Target & 1,823 & 97,230,848 & 97,419,200,814 & 34,700 \\
\bottomrule
\end{tabular}
\end{table}

\paragraph{Model and training.} The starting model is an approximately 185M-parameter decoder with a Gemma~3-style architecture~\citep{gemmateam2025gemma3}: RMSNorm, SwiGLU, RoPE, and hybrid local/global attention.
The model uses the Llama 3.1 tokenizer and a context length of 1,024. Training uses eight H200 GPUs, a micro-batch size of 352, and no gradient accumulation, giving approximately 2.88M tokens per step.
Each candidate is pretrained from scratch.
The training-step count scales with data volume to approximate one pass through each subset; baseline hyperparameters are specified per instance in Table~\ref{tab:pretraining_hpo_config}.

\paragraph{Editable interface.}
The proposer can modify four components:
\begin{itemize}\raggedright
\item Model architecture, including attention, normalization, and feed-forward computation.
\item Muon optimizer geometry~\citep{jordan2024muon}, including Newton--Schulz updates, clipping, and parameter grouping.
\item The Warmup--Stable--Decay learning-rate schedule.
\item Training mechanisms within the fixed data and compute budget.
\end{itemize}
Dataset assignment, tokenizer, sequence length, batch size, total steps, checkpoint isolation, and evaluation are protected.

\paragraph{Evaluator.} We use lm-evaluation-harness~\citep{gao2023lmeval} in zero-shot mode.
ARC-Challenge and ARC-Easy~\citep{clark2018arc}, HellaSwag~\citep{zellers2019hellaswag}, and OpenBookQA~\citep{mihaylov2018openbookqa} use normalized accuracy. PIQA~\citep{bisk2019piqa}, WinoGrande~\citep{sakaguchi2019winogrande}, and the OpenAI variant of LAMBADA~\citep{paperno2016lambada} use standard accuracy.
The evaluator loads the final checkpoint and takes the unweighted mean of these seven scores.
All seven evaluations must return successfully, and the implementation must pass the causal-leakage check, for a run to be valid.
Decomposed PE excludes full-target outcomes from adaptive proposal feedback; direct full-target evolution uses those outcomes during search.

\paragraph{Publication and promotion.} Published candidates must pass peer screening and both intermediate evaluations before full-target training.
Table~\ref{tab:pe_gates} specifies the improvement thresholds at each stage and the full-target graduation criterion.

\subsection{VLM pretraining}
\label{app:vlm}

\paragraph{Data hierarchy.} The benchmark uses a 1,000-shard image--conversation corpus in FineVisionMax format~\citep{wiedmann2025finevision}.
Shards 980--999 define the validation split; the full target uses the remaining 980 training shards.
Table~\ref{tab:vlm_data} gives the source, intermediate, and full-target inventories.

\begin{table}[htbp]
\centering\small
\setlength{\tabcolsep}{4pt}
\caption{VLM training hierarchy.
Rows report the raw training-data inventory and prescribed training steps.}
\label{tab:vlm_data}
\begin{tabular}{@{}llrrrr@{}}
\toprule
Dataset & Role & Shards & Shard indices & Raw rows & Steps \\
\midrule
D2-A & Source & 20 & 0--19 & 48,500 & 48 \\
D2-B & Source & 20 & 50--69 & 48,500 & 48 \\
D5-A & Source & 50 & 0--49 & 121,250 & 118 \\
D5-B & Source & 50 & 50--99 & 121,250 & 118 \\
D5-C & Source & 50 & 100--149 & 121,250 & 118 \\
D10 clean & Intermediate & 100 & 150--249 & 242,500 & 237 \\
D10+C & Intermediate & 100 & 100--149; 250--299 & 242,500 & 237 \\
Full target & Target & 980 & 0--979 & 2,376,500 & 2,320 \\
\bottomrule
\end{tabular}
\end{table}

\paragraph{Model and training.} The language backbone is a custom 28-layer hybrid decoder with a Qwen3.5-style architecture~\citep{qwenteam2026qwen35}, including full grouped-query attention and causal-convolutional linear attention, RMSNorm, SwiGLU, and RoPE.
A frozen SigLIP-so400m-patch14-384 encoder~\citep{zhai2023siglip} supplies 729 image-token positions per image; a trainable two-layer MLP projects these features into the language model.
The model has approximately 641.6M parameters in the language backbone, 646.1M trainable parameters including the projector, and 1.074B total parameters including the frozen vision encoder.

Training uses eight H200 GPUs, a micro-batch size of 8, gradient accumulation over 16 steps, a sequence length of 2,048, and BF16 precision.
Step counts scale with subset size to approximate one training epoch, ranging from 48 to 2,320.
Each candidate trains the language backbone and projector from scratch with the frozen vision encoder; evolved candidates do not inherit learned weights from earlier iterations.

\paragraph{Editable interface.}
The proposer can modify three components:
\begin{itemize}\raggedright
\item The language backbone's attention, normalization, RoPE, and SwiGLU blocks, and the MLP projector.
\item Optimizer updates, orthogonalization, Q/K clipping, and parameter grouping.
\item The Warmup--Stable--Decay learning-rate schedule.
\end{itemize}
The frozen SigLIP encoder, projector width, image-token interface, outer fusion/data path, data assignment, sequence length, batch and accumulation, steps, and evaluator remain fixed.

\paragraph{Evaluation metric and validity.}
The evaluator measures token-weighted next-token cross-entropy on validation data disjoint from training.
Before scoring, the evaluator verifies the frozen vision encoder, parameter budget, and causal validity.
The shifted next-token objective predicts position $t+1$ from position $t$; masked positions are excluded from the supervised loss.
The reported loss sums cross-entropy over these supervised tokens and divides by their total count, rather than averaging example losses.

\paragraph{Graduation.}
Candidates advance after peer screening and must pass both intermediate evaluations before full-target training.
Table~\ref{tab:pe_gates} specifies the relative and absolute loss-reduction criteria at each stage.
Decomposed PE withholds full-target outcomes from adaptive search feedback.

\subsection{Shared task guidance}
\label{app:environment-preparation}

Preliminary experiments identify recurring implementation errors, which we summarize in shared domain guidance.
All compared discovery methods receive the same guidance for a given task family in every LLM call.
Table~\ref{tab:domain-pitfalls} summarizes the guidance; Appendix~\ref{app:agent-prompts} shows how it accompanies task instructions and shared evidence.

\begin{table}[htbp]
\centering\small
\setlength{\tabcolsep}{4pt}
\renewcommand{\arraystretch}{1.12}
\caption{Shared domain guidance for valid candidate designs.
These instructions complement the fixed interfaces and evaluators.}
\label{tab:domain-pitfalls}
\begin{tabular*}{\linewidth}{@{\extracolsep{\fill}}P{1.8cm}P{11.1cm}@{}}
\toprule
Domain & Scientific and computational constraints \\
\midrule
Ranking & Avoid attention whose cost is quadratic in the full sequence length. \\
\addlinespace[3pt]
Watch-time & Preserve non-negative distributional support and stable likelihoods; treat CIKM16 duration as session context. \\
\addlinespace[3pt]
LLM & Distinguish vector and matrix parameters in Muon and maintain stable optimizer updates. \\
\addlinespace[3pt]
VLM & Preserve causal attention and the fixed visual interface. \\
\addlinespace[3pt]
RL & Preserve response groups and masks, and compute stable probability ratios. \\
\bottomrule
\end{tabular*}
\end{table}

\subsection{Baseline calibration and per-candidate hyperparameter optimization}
\label{app:baseline-construction}

\paragraph{Baseline calibration before discovery.}
Baseline HPO tunes the unmodified starting implementation separately for each instance before evolutionary search.
Ranking, watch-time, and LLM/VLM source and intermediate references use CMA-ES sampling in Optuna~\citep{akiba2019optuna}, with LLM-proposed configurations added after warm-up when the trial budget permits.
Results from all evaluated configurations update the CMA-ES search.
Tables~\ref{tab:baseline_calibration} and~\ref{tab:baseline_hpo_space} specify the calibration settings, trial caps, and search spaces.
The starting code, model capacity, data, evaluator, and prescribed epochs or step count remain fixed during calibration; ranking additionally searches the per-GPU batch size.
The selected configuration and measured reference score are then fixed and shared by competing discovery procedures in that instance.
RL disables baseline HPO and uses the mean of three independent fixed-configuration runs in each source and larger-scale instance.
LLM and VLM full targets each use one measured reference under a prescribed configuration shared by PE and Direct. The cost of full-target training precludes a separate full-target HPO sweep within the study budget; both methods use the same full-target reference and evaluation protocol.

\begin{table}[htbp]
\centering\small
\setlength{\tabcolsep}{5pt}
\renewcommand{\arraystretch}{1.1}
\caption{Baseline HPO trial caps, applied separately to each instance. Calibration also evaluates the original configuration and confirms the selected configuration. Early stopping uses a patience of ten trials without improvement.}
\label{tab:baseline_calibration}
\begin{tabular*}{\linewidth}{@{\extracolsep{\fill}}llr@{}}
\toprule
Family & Instances & Trial cap $N$ \\
\midrule
Ranking & Five sources and KuaiRand-27K & 50 \\
Watch-time & Five sources and KuaiRand-27K & 50 \\
LLM pretraining & Two 2\% sources and two intermediate instances & 20 \\
LLM pretraining & Three 5\% sources & 10 \\
VLM pretraining & Five sources and two intermediate instances & 20 \\
\bottomrule
\end{tabular*}
\par\smallskip
\begin{minipage}{\linewidth}\footnotesize
CMA-ES supplies the first ten warm-up trials; approximately 30\% of subsequent proposals come from the LLM when trials remain.
\end{minipage}
\end{table}

\begin{table}[htbp]
\centering\footnotesize
\setlength{\tabcolsep}{4pt}
\renewcommand{\arraystretch}{1.1}
\caption{Baseline HPO search spaces.
``Log'' denotes a logarithmic search scale; ``log integer'' additionally restricts values to integers.
Unmarked intervals use a linear scale.
LLM/VLM full-target configurations are fixed without HPO, and RL has no HPO search space.}
\label{tab:baseline_hpo_space}
\begin{tabular}{@{}P{2.1cm}P{10.9cm}@{}}
\toprule
Family / tier & Parameters and ranges \\
\midrule
Ranking & Dropout $[0.1,0.6]$; learning rate $[10^{-4},10^{-2}]$ (log); per-GPU batch size $[32,128]$ (log integer); weight decay $[10^{-5},10^{-2}]$ (log). \\
\addlinespace[3pt]
Watch-time & Learning rate $[0.01,0.3]$ (log); weight decay $[10^{-7},10^{-3}]$ (log); $\alpha\in[0,1]$; $\beta\in[0.1,5]$; dropout $[0,0.5]$. \\
\addlinespace[3pt]
LLM: source, intermediate & Peak learning rate $[5\times10^{-5},10^{-3}]$ (log); weight decay $[0.01,0.5]$ (log); warm-up steps $[50,300]$ (log integer); decay fraction $[0.05,0.3]$; Muon momentum $[0.9,0.99]$; Q/K clipping threshold $[50,200]$ (log). \\
\addlinespace[3pt]
VLM: source, intermediate & Same ranges as LLM except warm-up steps: $[2,12]$ for 2\% sources, $[5,30]$ for 5\% sources, and $[10,60]$ for 10\% intermediate instances (log integer). \\
\bottomrule
\end{tabular}
\end{table}

\paragraph{Selected configurations.}
Table~\ref{tab:ranking_hpo} gives the selected ranking hyperparameters, and Table~\ref{tab:watch_hpo} lists the six watch-time configurations.
Table~\ref{tab:pretraining_hpo_config} gives the LLM/VLM source, intermediate, and full-target configurations.
The fixed RL configuration is specified in Appendix~\ref{app:rl}.

\begin{table}[htbp]
\centering\footnotesize
\setlength{\tabcolsep}{5pt}
\caption{Selected ranking baseline hyperparameters.
Batch size is per GPU.
Learning rate and weight decay use the column scale factors.}
\label{tab:ranking_hpo}
\begin{tabular*}{\linewidth}{@{\extracolsep{\fill}}lrrrr@{}}
\toprule
Dataset & Dropout & \shortstack{Learning rate\\($\times10^{-3}$)} & Batch size & \shortstack{Weight decay\\($\times10^{-4}$)} \\
\midrule
KuaiLive & 0.49 & 2.5 & 47 & 4.8 \\
KuaiRand-1K & 0.352 & 3 & 38 & 1.1 \\
KuaiRand-Pure & 0.28727 & 7.969455 & 88 & 6.251374 \\
KuaiRec & 0.3 & 4.3 & 40 & 14 \\
KuaiSAR & 0.53 & 5.2 & 46 & 9 \\
\midrule
KuaiRand-27K & 0.378871 & 3.548962 & 51 & 1.985122 \\
\bottomrule
\end{tabular*}
\end{table}

\begin{table}[!ht]
\centering\footnotesize
\setlength{\tabcolsep}{3pt}
\renewcommand{\arraystretch}{1.08}
\caption{Fixed LLM/VLM baseline hyperparameters.
Learning rates use the indicated scale factor.
Columns give peak learning rate $\eta$, weight decay $\lambda$, warm-up steps $n_{\mathrm{warm}}$, decay fraction $f_{\mathrm{decay}}$, Muon momentum $\mu$, and Q/K clipping threshold $\tau_{\mathrm{QK}}$.
Full-target rows are prescribed configurations without HPO.}
\label{tab:pretraining_hpo_config}
\begin{tabular*}{\linewidth}{@{\extracolsep{\fill}}lrrrrrr@{}}
\toprule
Instance & \shortstack{$\eta$\\($\times10^{-4}$)} & $\lambda$ & $n_{\mathrm{warm}}$ & $f_{\mathrm{decay}}$ & $\mu$ & $\tau_{\mathrm{QK}}$ \\
\midrule
\multicolumn{7}{@{}l}{\textit{LLM pretraining}} \\
D2-A & 4.480393 & 0.018409 & 66 & 0.143635 & 0.985564 & 114.656 \\
D2-B & 5.4 & 0.05 & 170 & 0.085 & 0.919 & 170 \\
D5-A & 4.72 & 0.022 & 78 & 0.108 & 0.982 & 121 \\
D5-B & 4.480393 & 0.018409 & 66 & 0.143635 & 0.985564 & 114.656 \\
D5-C & 4.480393 & 0.018409 & 66 & 0.143635 & 0.985564 & 114.656 \\
D10 clean & 4.480393 & 0.018409 & 66 & 0.143635 & 0.985564 & 114.656 \\
D10+C & 5.2 & 0.055 & 172 & 0.118 & 0.9235 & 156 \\
\addlinespace[2pt]
Full target & 2 & 0.1 & 500 & 0.1 & 0.95 & 100 \\
\midrule
\multicolumn{7}{@{}l}{\textit{VLM pretraining}} \\
D2-A & 7.8 & 0.075 & 6 & 0.248 & 0.9165 & 168 \\
D2-B & 9.085181 & 0.114728 & 5 & 0.172398 & 0.913847 & 149.21 \\
D5-A & 7.2 & 0.035 & 20 & 0.218 & 0.918 & 185 \\
D5-B & 9 & 0.12 & 15 & 0.185 & 0.922 & 135 \\
D5-C & 4.55 & 0.073 & 18 & 0.203 & 0.929 & 112 \\
D10 clean & 5.178996 & 0.020461 & 27 & 0.164448 & 0.906785 & 119.578 \\
D10+C & 5.6 & 0.055 & 32 & 0.205 & 0.921 & 158 \\
\addlinespace[2pt]
Full target & 1 & 0.1 & 100 & 0.1 & 0.95 & 100 \\
\bottomrule
\end{tabular*}
\end{table}

\paragraph{Per-candidate HPO in the reported configuration.}
Per-candidate HPO adapts training hyperparameters for a proposed intervention during discovery and is separate from baseline calibration.
Per-candidate HPO is enabled only for KuaiRand-1K ranking, using the same procedure for PE and standalone evolution.
We enable it on this instance because performance is particularly sensitive to training hyperparameters and the small dataset makes repeated tuning affordable.
Table~\ref{tab:per_idea_hpo} records the reported GPT-5.5 settings, including the extra trials and optional paired baseline control.

\begin{table}[!ht]
\centering\footnotesize
\setlength{\tabcolsep}{4pt}
\renewcommand{\arraystretch}{1.1}
\caption{Per-candidate HPO in the reported GPT-5.5 configuration.
Counts cover one local mutation before any separate peer or promotion evaluations.
Disabled settings execute neither per-candidate HPO trials nor paired baseline controls.}
\label{tab:per_idea_hpo}
\begin{tabular*}{\linewidth}{@{\extracolsep{\fill}}lcrrr@{}}
\toprule
Family / sources & Enabled & Extra trials & Paired control & Total evaluations \\
\midrule
Ranking / KuaiRand-1K & Yes & 3 & Up to 1 & Up to 5 \\
Ranking / other four sources & No & 0 & 0 & 1 \\
Watch-time & No & 0 & 0 & 1 \\
RL algorithm & No & 0 & 0 & 1 \\
LLM pretraining & No & 0 & 0 & 1 \\
VLM pretraining & No & 0 & 0 & 1 \\
\bottomrule
\end{tabular*}
\end{table}

\paragraph{Paired baseline control and selection.}
After a successful main evaluation, per-candidate HPO starts from the receiving instance's calibrated baseline hyperparameters.
When a paired control is used, it evaluates the unmodified code under the candidate's selected hyperparameters, yielding $\widetilde b_e(h)$ for candidate $h$ on instance $e$.
This control measures the code intervention's effect at the candidate's selected hyperparameters.
Source publication retains the ordinary source rule in Table~\ref{tab:pe_gates} and, when a valid paired control exists, additionally requires the candidate to exceed that control by the prescribed relative improvement margin.
Cross-instance selection uses the stronger of the fixed reference and any valid paired control:
\begin{equation}
\label{eq:hpo-control}
s_e b_e^{\mathrm{gate}}(h)=\max\{s_e b_e,\,s_e\widetilde b_e(h)\}.
\end{equation}
When no valid paired control is available, we set $b_e^{\mathrm{gate}}(h)=b_e$.
Thus $b_e^{\mathrm{gate}}(h)$ is the larger reference for reward/accuracy and the smaller reference for error/loss.
Reported gains still use the fixed reference $b_e$ (Equation~\ref{eq:rmd-gain}); cross-instance selection uses $b_e^{\mathrm{gate}}(h)$.

\subsection{Discovery compute and search depth}
\label{app:benchmark-runtime}

We report compute for a complete discovery run, comprising five parallel searches and their prescribed evaluations.
Due to policy restrictions, we cannot disclose exact GPU-hour usage; Table~\ref{tab:discovery-compute-scales} reports the permitted orders of magnitude.
LLM and VLM pretraining require thousands of GPU-hours per complete run, ranking and RL algorithm discovery require hundreds, and watch-time prediction requires tens.
The benchmark and search code will be made publicly available to support reproduction under the stated protocols.

\begin{table}[!ht]
\centering\small
\setlength{\tabcolsep}{8pt}
\renewcommand{\arraystretch}{1.08}
\caption{\textbf{Compute scale of a complete discovery run.} Each run comprises five searches and their prescribed evaluations. Entries give orders of magnitude for total allocated GPU-hours; exact usage is not disclosed due to policy restrictions.}
\label{tab:discovery-compute-scales}
\begin{tabular}{@{}ll@{}}
\toprule
Task family & GPU-hours per complete discovery run \\
\midrule
Ranking & Hundreds \\
Watch-time prediction & Tens \\
RL algorithm discovery & Hundreds \\
LLM pretraining & Thousands \\
VLM pretraining & Thousands \\
\bottomrule
\end{tabular}
\end{table}

PE uses smaller training subsets for ongoing proposal--evaluation cycles and reserves full-target training for candidates that pass the peer and intermediate gates.
Appendix~\ref{app:execution-accounting} specifies the completed-run comparison and accounts for all executed search and evaluation stages.

\FloatBarrier

\section{Population Evolution Implementation and Reporting}
\label{app:implementation}

This appendix connects the search procedure in Section~\ref{sec:method} to its implementation.
We specify publication and promotion decisions, the evidence supplied to each source search, and asynchronous execution, then define the units used to report compute and outcomes.

\subsection{Publication and promotion thresholds}
\label{app:selection-thresholds}

PE tests publication and promotion margins against the reference score of the instance being evaluated. Based on preliminary studies, we set domain-specific margins to require challenging but attainable improvements under the prescribed training allocations and fix them for the reported comparisons. These margins define operational promotion and success criteria; Tables~\ref{tab:pe_joint} and~\ref{tab:pe_full} also report continuous performance gains.
Source publication requires the configured gain over the fixed reference and, when a valid paired control is available, the additional gain over that control (Appendix~\ref{app:baseline-construction}).
Cross-instance selection uses the gating reference $b_e^{\mathrm{gate}}(h)$ defined in Equation~\ref{eq:hpo-control}.
With $s_e$ defined in Equation~\ref{eq:rmd-gain}, the directed absolute gain is $a_e^{\mathrm{gate}}(h)=s_e\bigl(y_e(h)-b_e^{\mathrm{gate}}(h)\bigr)$ and the relative gain is $r_e^{\mathrm{gate}}(h)=a_e^{\mathrm{gate}}(h)/\max(|b_e^{\mathrm{gate}}(h)|,\epsilon)$.
An evaluation passes only when the run is valid, the score is finite, and every configured relative and absolute margin is met.
For an intervention $h$ originating at $E_i$, let $\rho_x$ and $\alpha_x$ denote the configured relative and absolute peer margins, and let $K_x$ be the required number of passing peers.
Counting only completed valid peer evaluations, the selector implements
\begin{equation}
\operatorname{eligible}(h)=
\mathbf 1\!\left[\sum_{E_j\in\mathcal D\setminus\{E_i\}}
\mathbf 1\!\left[r_j^{\mathrm{gate}}(h)\geq\rho_x\ \land\ a_j^{\mathrm{gate}}(h)\geq\alpha_x\right]\geq K_x\right].
\label{eq:cross-gate}
\end{equation}
An omitted margin imposes no constraint; in our experiments, $K_x=2$ for every domain.
Scores are not averaged across instances before selection.
An eligible candidate proceeds to the larger-scale instance in Track~A or to the two intermediate instances in Track~B.
Both intermediate instances must pass before full-target training.
Graduation then requires the Track~A larger-scale or Track~B full-target margin.
Table~\ref{tab:pe_gates} specifies the margins used by PE; the benchmark's training and evaluation instances can also support other selection policies.

\begin{table}[htbp]
\centering\small
\setlength{\tabcolsep}{3pt}
\renewcommand{\arraystretch}{1.1}
\caption{PE's publication and promotion margins.
Relative gains are direction-normalized; where an absolute margin is listed, both conditions apply.
Every peer gate requires at least two wins among four peers.}
\label{tab:pe_gates}
\begin{tabular}{@{}P{1.7cm}P{2.55cm}P{2.65cm}P{2.95cm}P{2.6cm}@{}}
\toprule
Family & Source publication & Peer evaluation & Larger scale / intermediate & Full target \\
\midrule
Ranking & 2\% relative & 1\% relative & 2\% relative & --- \\
Watch-time & 0.5\% relative & 0.5\% relative & 0.5\% relative & --- \\
RL algorithm & 10\% rel.; $1.5/408$ abs. & 5\% rel.; $0.5/408$ abs. & 4\% rel.; $4/423$ abs. & --- \\
LLM & 2\% relative & 1\% relative & 2\% on both intermediates & 1\% relative \\
VLM & 10\% relative & 10\% rel.; 0.02 abs. & 6\% rel.; 0.02 abs., both intermediates & 4\% rel.; 0.015 abs. \\
\bottomrule
\end{tabular}
\par\vspace{3pt}
\begin{minipage}{\linewidth}\footnotesize
Graduation requires meeting the larger-scale criteria in Track~A, or the full-target criteria after both intermediate evaluations pass in Track~B.
\end{minipage}
\end{table}

\subsection{Shared candidate records and search context}
\label{app:memory-projection}

The shared ledger connects candidate evaluation to ongoing local search.
A source publishes a promising intervention, peers evaluate it, and the accumulated results enter later proposal prompts (Figure~\ref{fig:pe_overview}(a)).
The following describes this information flow; Appendix~\ref{app:agent-prompts} shows how it enters proposal prompts and candidate construction.

\paragraph{Publishing a candidate.}
After a candidate meets its source instance's publication margin (Appendix~\ref{app:selection-thresholds}), PE creates a shared record containing the executable intervention and a short description of its mechanism.
The record identifies the source instance and iteration, reports its score and reference baseline, and tracks the candidate's evaluation results and current status.
The record keeps executable code distinct from the mechanism description: peer evaluation executes the code, while proposal context summarizes the mechanism and its measured outcomes.

\paragraph{Evaluating the intervention on peers.}
For each peer, the coordinator restores that instance's reference implementation, applies the published intervention, and runs the peer's prescribed training and evaluation.
Model parameters are trained separately under the receiving instance's protocol.
This \emph{cross-instance replay} measures where the published intervention remains useful.
Each completed evaluation adds the receiving instance, score, gain, and pass decision to the candidate's record.
Valid weak or near-neutral results remain available even when they do not meet a promotion margin.
The completed peer results determine eligibility for larger-scale or intermediate evaluation under the criteria in Appendix~\ref{app:selection-thresholds}.

\begin{samepage}
\paragraph{Building context for the next local proposal.}
Before each LLM call on source instance $E_i$, PE selects relevant records from the ledger to form the shared context $\Phi_i(M_t)$ in Equation~\ref{eq:proposal-policy}.
This context contains three groups:
\begin{itemize}
\item \textbf{Own candidates:} at most ten candidates published by $E_i$, with available peer and larger-scale or intermediate results.
\item \textbf{Successful peer transfers:} at most six candidates from other source instances that succeeded on $E_i$, ranked by their measured benefit on $E_i$.
\item \textbf{Weak or unsuccessful peer transfers:} at most six candidates from other source instances that failed or transferred weakly on $E_i$.
\end{itemize}
\end{samepage}
Each selected entry summarizes the mechanism, originating instance, evaluation status, and measured effects; all entries together are limited to 4K tokens.
Gains are direction-normalized as in Equation~\ref{eq:rmd-gain}, so positive values indicate improvement for both accuracy and loss metrics.
The local harness retains its own program archive, while the proposer uses the shared evidence to reuse, modify, or combine mechanisms in later candidates.
This separates two uses of shared records: replay tests an existing intervention, while proposal context helps generate the next one.

\subsection{Prompts and local code edits}
\label{app:agent-prompts}

Each source instance uses a local search harness to select a parent program and ask an LLM for a code edit.
The main experiments use the AdaEvolve-based implementation in SkyDiscover~\citep{liu2026skydiscover,cemri2026adaevolve}.
At each proposal step, the prompt combines the local program and search history with the shared context described in Appendix~\ref{app:memory-projection}.

\paragraph{Inputs to the proposer.}
The template combines fixed task instructions and shared domain guidance with the selected parent, local history, and latest available shared evidence.

\begin{quote}\begin{minipage}{\linewidth}\footnotesize
\begin{verbatim}
SYSTEM
  Objective, dataset, metric, and fixed training budget
  Editable regions, interfaces, and fixed controls
  Shared context: own candidates and their transfer results
                  peer candidates evaluated on this instance
  Domain guidance and harness tool instructions

USER
  Selected parent code and metrics
  Previous attempts and evaluator feedback
  Program examples sampled by the harness
  Search objective and edit requirements
\end{verbatim}
\end{minipage}\end{quote}

\paragraph{Example of the shared context.}
Consider a proposal on VLM source instance D2-A.
Its shared context can include the candidate discovered on D5-C at iteration 55 and subsequently evaluated on D2-A.
The intervention replaces global Muon gradient clipping with clipping based on the final update's RMS, yielding relative gains of approximately 10.50\% on D5-C and 10.34\% on D2-A.

\begin{quote}\begin{minipage}{\linewidth}\footnotesize
\begin{verbatim}
Candidate from a peer instance
  Source: VLM D5-C, iteration 55
  Mechanism: final-update RMS trust clipping for Muon
  Source gain: +10.50%; gain on VLM D2-A: +10.34%
  Receiving-instance decision: pass
  Full-target outcomes: withheld from proposal context
\end{verbatim}
\end{minipage}\end{quote}

In this Track~B example, the proposer receives peer and intermediate evidence but no full-target scores or pass/fail outcomes (Figure~\ref{fig:pe_overview}(c)).
The task objectives, editable components, and fixed experimental controls follow Appendices~\ref{app:ranking}--\ref{app:vlm}.

\paragraph{Output: a local edit to the selected parent.}
The LLM edits the selected parent within the permitted components, preserving the specified interfaces and fixed experimental controls. The candidate is then trained and evaluated on its source instance.
The evaluation updates the local history and archive; candidates meeting the publication margin enter the shared ledger for cross-instance replay (Appendix~\ref{app:memory-projection}).

\subsection{Asynchronous discovery loop}
\label{app:algorithm}

Algorithm~\ref{alg:pe} summarizes the population operations.
Local searches and evaluation jobs proceed asynchronously.
Each source harness refreshes its shared-evidence context before every LLM call.
Iteration numbers are local to each source; timestamps determine which shared results are available to a proposal.

\begin{algorithm}[htbp]
\caption{Population Evolution: event-driven orchestration}
\label{alg:pe}
\small
\begin{algorithmic}[1]
\Require Source instances $\mathcal D$; local budgets; baselines; selector $\mathcal S$; track
\State Initialize separate local states $\tau_i$, shared ledger $M$, and evaluation queues
\While{local proposals or queued evaluations remain within the run allocation}
\If{the local search on source instance $E_i$ is ready for an iteration}
\State Run its proposal harness with $\tau_i$ and a per-call refresh of $\Phi_i(M)$
\State Obtain candidate $h$ and enqueue training and evaluation on $E_i$
\EndIf
\If{a local evaluation completes}
\State Update local history and code archive; publish if the source criterion is met
\State If published, queue evaluation of the same code change on peer instances
\EndIf
\If{a peer, intermediate, or Track~A larger-scale evaluation completes}
\State Record the job's outcome and pass flag; recompute eligibility using $\mathcal S$
\State If eligible, queue any unscheduled larger-scale (Track A) or intermediate (Track B) evaluations
\State In Track B, queue full-target training once, after both intermediate gates pass
\EndIf
\If{a full-target evaluation completes in Track B}
\State Record the full-target outcome and terminal state in $M$; exclude both from all proposal contexts
\EndIf
\EndWhile
\end{algorithmic}
\end{algorithm}

\subsection{Execution accounting and evidence units}
\label{app:execution-accounting}

\paragraph{Iterations, evaluations, and complete runs.}
A source iteration is one local proposal--evaluation cycle.
A complete discovery run comprises five source searches for PE or five independent searches for S/Direct, together with their prescribed transfer and target evaluations.
Discovery compute includes local search, peer replay, intermediate screening, and final-stage evaluation.
Per-candidate HPO trials, paired controls, retries, failed attempts, and re-evaluations are charged to discovery compute.
Shared baseline calibration is performed before search and accounted for separately.

\paragraph{Comparison budgets.}
Track~A matches source iterations; Track~B compares completed PE and Direct runs under matched total GPU compute across all stages.
For Track~A larger-scale evaluation, S selects the top $k=\lceil N_{\mathrm{PE}}/5\rceil$ candidates by source score from each of its five searches, where $N_{\mathrm{PE}}$ is the number of PE larger-scale candidate evaluations.
Rounding up gives S at least as many larger-scale candidate evaluations as PE.
Allocated compute sums GPU count multiplied by training--evaluation runtime across all jobs.
Exact GPU-hour usage is restricted by policy; Appendix~\ref{app:benchmark-runtime} reports the compute scales and code-release commitment.
Table~\ref{tab:pe_full} and both panels of Figure~\ref{fig:budget-discovery} summarize the completed runs; the LLM runs have 39 full-target attempts for PE and 70 candidate-mutation attempts for Direct.

\paragraph{Outcome summaries.}
For Track~A, ranking and watch-time average each source's best local gain across distinct datasets, whereas RL reports the best gain across five parallel searches of the same task (Table~\ref{tab:pe_source_detail}).
Peer-evaluation results and larger-scale outcomes are summarized separately.
Track~B reports the best full-target gain and the number of successful candidates from each completed run, using the criteria in Table~\ref{tab:pe_gates}.
Comparisons and relative gains are computed before display rounding.
Reported gains always use the fixed reference in Equation~\ref{eq:rmd-gain}; paired controls can strengthen selection without redefining this reference.
Appendix~\ref{app:vlm-checkpoint} gives raw attempt counts and passing means.

\paragraph{Replication.}
Given the substantial training cost of complete discovery runs, we report one complete run per method--configuration and allocate the available compute to comparisons across task families and model proposers.
The five searches are components of one complete run; repeated baseline evaluations estimate the reference score rather than variation across discovery runs.
For the configurations examined in preliminary independent runs, the relative performance ordering was consistent across runs.

\subsection{Full-target outcome details}
\label{app:vlm-checkpoint}

Each candidate contributes at most one success, regardless of repeated evaluations.

\paragraph{LLM outcomes.}
Over its completed run, PE evaluates 39 candidates in 39 full-target attempts; 34 return numeric scores and five return no score.
Thirteen meet the success criterion, with a mean gain of 1.67\% among the successful candidates.
Attempts without a score remain in the attempt count and do not contribute a success.
Direct produces no successful candidates from 70 candidate-mutation attempts; this count excludes the five initialization evaluations.
Both methods' outcomes are reported after their complete discovery runs under matched total GPU compute (Appendix~\ref{app:execution-accounting}).
Appendix~\ref{app:budget-results} varies the full-target success threshold using these same results.

\paragraph{VLM outcomes.}
Direct completes 100 attempts across five searches, with 96 finite scores and four failed attempts.
Seventeen evaluations meet the full-target criterion, corresponding to 12 successful candidates.
Relative to the fixed baseline loss 1.2071, the best loss is 1.1461 (5.05\% reduction) and the passing-evaluation mean is 1.1535 (4.44\% reduction).
PE has 61 full-target attempts, 57 finite scores, and four failed attempts.
All 57 finite outcomes meet the criterion, yielding 57 successful candidates with a mean gain of 7.66\%.

\section{Per-instance Discovery Outcomes}
\label{app:additional-details}
\label{app:source-outcomes}

Table~\ref{tab:pe_source_detail} reports local discovery outcomes for S and PE alongside PE's best gains from local and replayed peer candidates (Section~\ref{sec:transfer-analysis}).
For ranking and watch-time, the reported means average the best local gain on each dataset.
The five RL searches share the same base model and training data and differ only in random seed. Their goal is to find the strongest algorithm for this task, so we report the best result across the five searches.

\begin{table}[!ht]
\centering
\small
\setlength{\tabcolsep}{5pt}
\caption{Local discovery outcomes and PE's best gains from local and replayed peer candidates.
Gains are relative improvements.
For recommendation, S and PE report the best locally proposed gain on each dataset; the RL row reports the maximum across five searches of the same task.
Watch-time uses the GPT-5.5 paired comparison.}
\label{tab:pe_source_detail}
\begin{tabular}{@{}llrrr@{}}
\toprule
Domain & Dataset / summary & S & PE & \shortstack{PE best including\\peer candidates}$^{\dagger}$ \\
\midrule
Ranking & KuaiLive & +8.93\% & +23.19\% & +23.19\% \\
 & KuaiRand-1K & +5.53\% & +9.29\% & +10.66\% \\
 & KuaiRand-Pure & +8.03\% & +6.07\% & +6.07\% \\
 & KuaiRec & +1.03\% & +2.42\% & +6.47\% \\
 & KuaiSAR & +11.52\% & +3.88\% & +10.06\% \\
\cmidrule(lr){2-5}
 & \textbf{Mean} & +7.01\% & \textbf{+8.97\%} & +11.29\% \\
\midrule
Watch-time & CIKM16 & +7.32\% & +12.55\% & +12.55\% \\
 & KuaiRec & +0.40\% & +0.67\% & +0.67\% \\
 & WeChat & +0.80\% & +0.81\% & +0.81\% \\
 & KuaiRand-Pure & +0.75\% & +0.16\% & +0.65\% \\
 & KuaiRand-1K & +4.93\% & +5.08\% & +5.08\% \\
\cmidrule(lr){2-5}
 & \textbf{Mean} & +2.84\% & \textbf{+3.85\%} & +3.95\% \\
\midrule
RL algorithm & \textbf{Max / 5 searches} & +23.53\% & \textbf{+32.35\%} & --- \\
\bottomrule
\end{tabular}
\par\vspace{4pt}
\begin{minipage}{\linewidth}
\footnotesize
S: standalone evolution.
$^{\dagger}$The final column takes the best valid score among PE's locally proposed candidates and candidates from peer instances retrained on the receiving source instance.
It summarizes the value of peer evaluation within PE.
Comparisons use unrounded values.
\end{minipage}
\end{table}

\paragraph{Variation across datasets.}
Table~\ref{tab:source-paired-differences} reports the PE--S difference on each recommendation source, separately for locally proposed candidates and for PE's best available candidates including peer replay.
PE's local gains exceed S on three of five ranking datasets and four of five watch-time datasets, with the largest differences on KuaiLive and CIKM16.
Omitting KuaiLive changes the ranking local-mean difference to approximately $-1.11$ percentage points; omitting CIKM16 changes the watch-time difference to approximately $-0.04$ points.
When PE's replayed peer candidates are included, the corresponding differences remain positive at approximately $+1.79$ and $+0.08$ points, and the mean difference remains positive after omitting any one dataset in either family.
Thus local-search benefits are uneven, while direct reuse broadens the available improvements. These are descriptive aggregation checks derived from the displayed source results.

\begin{table}[!htbp]
\centering\small
\setlength{\tabcolsep}{6pt}
\renewcommand{\arraystretch}{1.1}
\caption{\textbf{Dataset-wise PE--S differences in relative gain.} Entries are percentage-point differences derived from the rounded values in Table~\ref{tab:pe_source_detail}. The two columns distinguish local proposals from the best PE candidates available after peer replay.}
\label{tab:source-paired-differences}
\begin{tabular*}{\linewidth}{@{\extracolsep{\fill}}llrr@{}}
\toprule
Domain & Dataset & Local PE minus S & PE including peers minus S \\
\midrule
Ranking & KuaiLive & +14.26 & +14.26 \\
 & KuaiRand-1K & +3.76 & +5.13 \\
 & KuaiRand-Pure & $-1.96$ & $-1.96$ \\
 & KuaiRec & +1.39 & +5.44 \\
 & KuaiSAR & $-7.64$ & $-1.46$ \\
\midrule
Watch-time & CIKM16 & +5.23 & +5.23 \\
 & KuaiRec & +0.27 & +0.27 \\
 & WeChat & +0.01 & +0.01 \\
 & KuaiRand-Pure & $-0.59$ & $-0.10$ \\
 & KuaiRand-1K & +0.15 & +0.15 \\
\bottomrule
\end{tabular*}
\end{table}
\FloatBarrier

\section{Graduated Designs and Transfer Analysis}
\label{app:mechanism-analysis}

Table~\ref{tab:graduated-designs} illustrates one graduated design from each task family. Tables~\ref{tab:pe_joint} and~\ref{tab:pe_full} report the strongest outcome in each family.
The 13 successful LLM candidates span three intervention families: Muon update geometry, RoPE configurations within the existing hybrid attention stack, and late-training EMA/SWA checkpoint averaging.
\begin{table}[htbp]
\centering\footnotesize
\setlength{\tabcolsep}{4pt}
\renewcommand{\arraystretch}{1.15}
\caption{\textbf{One concrete graduated design from each task family.}
Gains are relative to the fixed reference at the stated final-stage instance.
Each row identifies a complete candidate; the intervention column summarizes its principal changes.
The success criteria are specified in Table~\ref{tab:pe_gates}.}
\label{tab:graduated-designs}
\begin{tabular}{@{}P{1.5cm}P{2.3cm}P{5.3cm}P{3.0cm}@{}}
\toprule
Family & Source / iteration & Principal intervention & Final-stage outcome \\
\midrule
Ranking & KuaiRand-Pure / 57 & Gated length-64 causal attention alongside linear retention; inherited temporal and spectral components. & KuaiRand-27K: 3.02\% ranking gain. \\
\addlinespace[4pt]
Watch-time & WeChat / 21 & Duration-conditioned, left-truncated Laplace mixture with a learned near-median quantile readout. & KuaiRand-27K: 2.17\% MAE reduction. \\
\addlinespace[4pt]
RL algorithm & Source 2 / 87 & Bounded advantage weights based on trajectory distance from the reference policy. & 3B, 50 steps: 6.64\% pass@1 gain. \\
\addlinespace[4pt]
LLM & D5-B / 89 & EMA/SWA checkpoint averaging over approximately the final 20\% of training. & Full target: 2.48\% accuracy gain. \\
\addlinespace[4pt]
VLM & D5-A / 86 & Gated global image-context injection into patch tokens, combined with square-root warm-up. & Full target: 7.51\% loss reduction. \\
\bottomrule
\end{tabular}
\end{table}

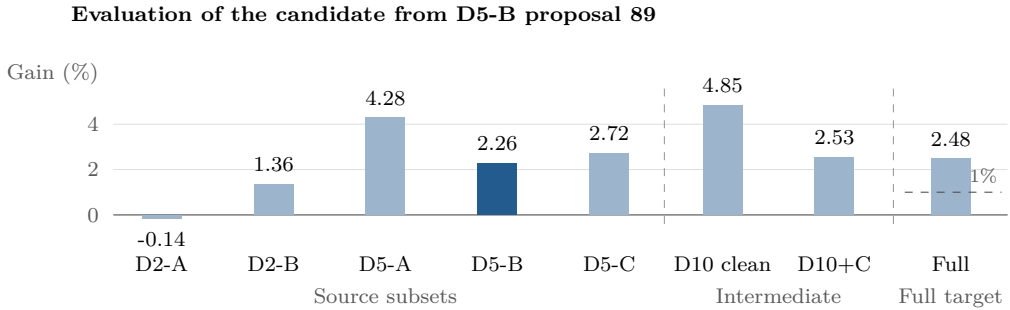
\begin{figure}[htbp]
\centering
\begin{tikzpicture}[x=1cm,y=1cm,font=\footnotesize]
\node[anchor=west,font=\footnotesize\bfseries] at (.05,1.48) {Evaluation of the candidate from D5-B proposal 89};
\begin{scope}[shift={(0,-1.15)}]
\node[anchor=east,text=black!65] at (.66,1.87) {Gain (\%)};
\draw[black!12,line width=.4pt] (.73,0.0000)--(12.56,0.0000);
\node[anchor=east,text=black!65] at (.66,0.0000) {0};
\draw[black!12,line width=.4pt] (.73,0.6000)--(12.56,0.6000);
\node[anchor=east,text=black!65] at (.66,0.6000) {2};
\draw[black!12,line width=.4pt] (.73,1.2000)--(12.56,1.2000);
\node[anchor=east,text=black!65] at (.66,1.2000) {4};
\draw[black!55,line width=.5pt] (.73,0)--(12.56,0);
\fill[peblue!45] (1.1100,0) rectangle (1.6300,-0.0429);
\node[anchor=north] at (1.3700,-0.1029) {-0.14};
\node[anchor=north] at (1.3700,-.42) {D2-A};
\fill[peblue!45] (2.5900,0) rectangle (3.1100,0.4074);
\node[anchor=south] at (2.8500,0.4574) {1.36};
\node[anchor=north] at (2.8500,-.42) {D2-B};
\fill[peblue!45] (4.0700,0) rectangle (4.5900,1.2837);
\node[anchor=south] at (4.3300,1.3337) {4.28};
\node[anchor=north] at (4.3300,-.42) {D5-A};
\fill[peblue] (5.5500,0) rectangle (6.0700,0.6792);
\node[anchor=south] at (5.8100,0.7292) {2.26};
\node[anchor=north] at (5.8100,-.42) {D5-B};
\fill[peblue!45] (7.0300,0) rectangle (7.5500,0.8157);
\node[anchor=south] at (7.2900,0.8657) {2.72};
\node[anchor=north] at (7.2900,-.42) {D5-C};
\fill[peblue!45] (8.5300,0) rectangle (9.0500,1.4544);
\node[anchor=south] at (8.7900,1.5044) {4.85};
\node[anchor=north] at (8.7900,-.42) {D10 clean};
\fill[peblue!45] (10.0100,0) rectangle (10.5300,0.7590);
\node[anchor=south] at (10.2700,0.8090) {2.53};
\node[anchor=north] at (10.2700,-.42) {D10+C};
\fill[peblue!45] (11.5500,0) rectangle (12.0700,0.7428);
\node[anchor=south] at (11.8100,0.7928) {2.48};
\node[anchor=north] at (11.8100,-.42) {Full};
\draw[black!40,dashed] (8.02,-.12)--(8.02,1.67);
\draw[black!40,dashed] (11.05,-.12)--(11.05,1.67);
\draw[black!70,dashed] (11.20,.30)--(12.48,.30);
\node[anchor=south east,text=black!65,font=\scriptsize] at (12.55,.30) {1\%};
\node[anchor=north,text=black!65] at (4.33,-.85) {Source subsets};
\node[anchor=north,text=black!65] at (9.53,-.85) {Intermediate};
\node[anchor=north,text=black!65] at (11.81,-.85) {Full target};
\end{scope}
\end{tikzpicture}
\caption{\textbf{The checkpoint-averaging candidate from D5-B proposal 89 improves seven of eight instances.}
Gains use each instance's fixed reference; the candidate exceeds the 1\% full-target threshold.
The darker bar marks its source, vertical dividers separate training stages, and the dashed horizontal line marks the full-target threshold.}
\label{fig:checkpoint-transfer}
\end{figure}

\paragraph{Interpreting the transfer profiles.}
Figure~\ref{fig:transfer-profiles} shows three individual PE candidates: (a) the ranking design discovered on KuaiRand-Pure at iteration 57, (b) a ranking design discovered on KuaiLive, and (c) the watch-time head discovered on WeChat at iteration 21.
The profiles measure transfer after retraining the same design under each receiving instance's protocol; unevaluated pairs remain unknown.
Figure~\ref{fig:idea-cases} traces the availability of shared evidence and subsequent local proposals. Figure~\ref{fig:checkpoint-transfer} evaluates the checkpoint-averaging design from D5-B iteration 89.

\subsection{Execution traces and mechanism analysis}
\label{app:collaboration-cases}

The cases use the runs underlying Tables~\ref{tab:pe_joint} and~\ref{tab:pe_full}, with archived candidate code, evaluation results, and proposal records supporting the mechanism analysis.
The shared-context reconstruction combines archived ledger records available before each proposal with the context-selection rule; saved base prompts do not include the added shared-context text.
The evaluations measure transfer, while the traces establish when evidence was available for reuse. Proposal indices are local to each source search; the sequence below follows event timing across asynchronous searches. Gains use the fixed instance reference unless a direct-parent comparison is stated.

\paragraph{Case 1: LLM checkpoint averaging and continued local refinement.}
\label{app:case-llm}
The D2-B iteration-100 candidate implements checkpoint averaging using EMA/SWA late in training, alongside inherited short-context RoPE and Muon update-order changes.
It improves three of four peers and gains 1.06\% on the full target, above the graduation threshold, while regressing by 1.12\% on D2-A.
The reconstructed context for D5-B proposal 59 includes this successful peer-evaluation record, which was available before the proposal.
D5-B's iteration-59 proposal adopts checkpoint averaging using EMA/SWA in a new local candidate, beginning averaging roughly 55--65\% into training or during the WSD decay phase.
The resulting design improves D5-B by 3.30\% but gains only 0.90\% on the full target, below the 1\% graduation threshold.
The later D5-B iteration-89 candidate averages checkpoints over approximately the final 20\% of training.
On the full target, this candidate reaches 41.25\% accuracy, compared with 41.13\% for its direct parent---a 0.28\% relative improvement computed from unrounded scores (Table~\ref{tab:case-parent-comparisons}). This improvement accompanies a 1.17\% source regression. Its 2.48\% full-target gain is relative to the fixed 40.25\% reference; iteration 59 is an earlier design in the search, not its direct parent.
Full-target outcomes are excluded from proposal feedback and are used here to compare the resulting designs.
Checkpoint averaging plausibly reduces sensitivity to the final optimizer iterate. The trace documents the peer evidence available before the local edits and the transfer performance of the resulting designs.

\begin{table}[htbp]
\centering\small
\setlength{\tabcolsep}{4pt}
\caption{\textbf{Direct-parent comparisons for the representative cases.} Positive relative gains indicate improvement and are computed from unrounded scores. These comparisons use the direct parent rather than the fixed benchmark reference.}
\label{tab:case-parent-comparisons}
\begin{tabular*}{\linewidth}{@{\extracolsep{\fill}}llrrr@{}}
\toprule
Design & Metric & Parent score & Candidate score & Gain (\%) \\
\midrule
LLM, D5-B / 89 & Full-target accuracy (\%) & 41.13 & 41.25 & 0.28 \\
VLM, D5-A / 86 & Full-target loss & 1.1200 & 1.1164 & 0.33 \\
RL, source 2 / 87 & Source pass@1 (\%) & 3.43 & 3.19 & $-7.14$ \\
\bottomrule
\end{tabular*}
\end{table}

\paragraph{Case 2: Watch-time design reuse across datasets.}\label{app:watch-reuse-case}
The WeChat iteration-21 watch-time head combines a duration-conditioned, left-truncated Laplace mixture with a learned near-median quantile readout.
The design conditions on $\log(1+\mathrm{duration})$ and softly anchors component locations to video duration.
The mixture models concentrated viewing modes and long watch times from repeated viewing, while the near-median readout aligns point prediction with absolute error.
A differentiable quantile is learned within approximately $[0.45,0.55]$, with a small mixture-dependent gate for an additional soft duration adjustment.
Bisection locates the quantile, and a Newton correction propagates gradients through the distribution parameters.
Relative to its parent, the combined edit reduces source MAE from 17.79 to 17.72 seconds, a 0.39\% local improvement.
When retrained unchanged, the complete design improves KuaiRand-Pure by 0.65\%, exceeding the 0.16\% gain of its best locally proposed PE candidate.
It achieves the best reported larger-scale result, a 2.17\% MAE reduction on KuaiRand-27K, while regressing by 2.70\% on CIKM16 (Figure~\ref{fig:transfer-profiles}c).
On VK-LSVD under shared target-side calibration, the same head reduces MAE by 1.02\% across three training seeds (Appendix~\ref{app:transfer-watchtime}).
This case shows how PE turns a local discovery into a design useful to other instances, with benefits that extend to a dataset outside discovery.

\paragraph{Ranking: a local transition channel improves receiving source instances.}
The KuaiRand-Pure iteration-57 candidate adds a length-64 strictly causal local attention branch alongside linear retention, combining them through a small learned gate.
Inherited temporal features and a causal spectral filter remain part of the complete design.
When retrained on KuaiRec and KuaiSAR, this design outperforms the best candidates proposed locally on those instances. It retains a 3.02\% gain on KuaiRand-27K (Figure~\ref{fig:transfer-profiles}).
The local attention branch models recent transitions alongside the existing long-range aggregation.
Peer evaluations establish transfer of the complete design; their results subsequently entered the shared context of ongoing peer searches.

\paragraph{RL: reference-distance weighting and selective transfer.}
The source-2 iteration-87 design computes each trajectory's masked mean absolute difference between rollout-policy and reference-policy log probabilities.
It centers the distance using the batch median, scales it by a robust spread, clamps it to $[-6,6]$, and applies a sigmoid.
After advantage whitening, bounded weights amplify positive advantages for reference-like trajectories and attenuate them for distant trajectories. This modulation is reversed for negative advantages.
The distance is detached, and the KL coefficient and training budget remain fixed.
This reference-dependent weighting is intended to retain useful coding priors while learning from reward.
The candidate regresses against its direct parent on the source (Table~\ref{tab:case-parent-comparisons}) but remains above the fixed source reference.
It passes on two peers and regresses on the other two, then achieves a 6.64\% larger-scale gain.
The 32.35\% source-local maximum instead comes from the source-2 iteration-88 quantile-threshold rollout filter; the local and larger-scale maxima are distinct discoveries.
Appendix~\ref{app:transfer-rl} separately tests policies trained with the discovered algorithms on held-out coding tasks.

\paragraph{VLM: convergent discovery of global image context.}
The D5-C iteration-56 design pools frozen SigLIP patch features and applies image-level FiLM to projector hidden states through a 0.1-scaled residual.
It reduces full-target loss by 7.07\% relative to the fixed baseline despite regressing against its direct parent on both the source and full target.
D5-A had explored global image context before this candidate was published, and the D5-C record was not among the successful peer transfers shown to the D5-A iteration-83 proposal.
The trace therefore supports convergent exploration of this mechanism family.
D5-A iteration 83 mean-pools all image tokens, projects the summary, and adds a small gated residual to each token.
At D5-A iteration 86, the broadcast intervention is applied to a different local parent; the resulting candidate uses square-root warm-up.
The resulting full-target loss improves over its parent (Table~\ref{tab:case-parent-comparisons}), while source loss increases by only 0.02\%.
Its loss reduction relative to the 1.2071 fixed reference is 7.51\%.
The design supplies explicit image-level context to every patch while preserving the fixed image-token interface.

\section{Additional Configuration and Budget Results}
\label{app:cross-implementations}

\subsection{Larger-scale and full-target outcomes}
\label{app:outcomes-overview}
\begin{figure}[htbp]
\centering
\begin{tikzpicture}[
  x=.78cm,y=1cm,
  font=\footnotesize,
  pe/.style={draw=peblue,fill=peblue},
  control/.style={draw=black!55,fill=white,line width=.75pt},
  val/.style={font=\footnotesize,inner sep=2pt},
  name/.style={anchor=east,inner sep=0pt},
  tier/.style={anchor=east,font=\footnotesize\bfseries,inner sep=0pt}
]
\draw[control] (2.2,.48) circle[radius=2.0pt];
\node[anchor=west,inner sep=4pt] at (2.2,.48) {S / Direct};
\fill[pe] (4.8,.48) circle[radius=2.2pt];
\node[anchor=west,inner sep=4pt] at (4.8,.48) {Population Evolution};
\foreach \tick in {0,2,4,6,8,10} {
  \draw[black!10,line width=.35pt] (\tick,.18) -- (\tick,-2.19);
  \node[anchor=north,font=\footnotesize,inner sep=2pt] at (\tick,-2.29) {\tick};
}
\draw[black!40,line width=.45pt] (0,-2.29) -- (10,-2.29);
\node[anchor=north,font=\footnotesize,inner sep=2pt] at (5,-2.57)
  {Relative improvement over the fixed reference (\%)};
\draw[black!15,line width=.4pt] (-5.9,-1.17) -- (10,-1.17);
\node[tier] at (-3.8,-.33) {Track A};
\node[anchor=east,font=\footnotesize,inner sep=0pt] at (-3.8,-.68) {Larger scale};
\node[tier] at (-3.8,-1.55) {Track B};
\node[anchor=east,font=\footnotesize,inner sep=0pt] at (-3.8,-1.90) {Full target};
\node[name] at (-.2,0) {Ranking};
\node[name] at (-.2,-.45) {Watch-time};
\node[name] at (-.2,-.9) {RL algorithm};
\node[name] at (-.2,-1.5) {LLM pretraining};
\node[name] at (-.2,-1.95) {VLM pretraining};
\foreach \row/\base/\pop/\blabel/\plabel in {
  0/4.70/6.19/4.70/6.19,
  -.45/1.774/2.174/1.77/2.17,
  -.9/1.95/6.64/1.95/6.64,
  -1.5/.92/2.48/0.92/2.48,
  -1.95/5.052/8.78/5.05/8.78
} {
  \draw[black!30,line width=.8pt] (\base,\row) -- (\pop,\row);
  \draw[control] (\base,\row) circle[radius=2.0pt];
  \fill[pe] (\pop,\row) circle[radius=2.2pt];
  \node[val,anchor=east,text=black!65] at (\base-.10,\row) {\blabel};
  \node[val,anchor=west,text=peblue] at (\pop+.10,\row) {\plabel};
}
\end{tikzpicture}
\caption{\textbf{PE improves the best larger-scale or full-target outcome in all five task families.}
Each pair compares PE with standalone evolution (Track~A) or direct full-target evolution (Track~B).
The Track~A values match Table~\ref{tab:pe_joint}D, and the Track~B values match Table~\ref{tab:pe_full}.
These are completed-run best gains; Track~B uses matched total GPU compute. Figure~\ref{fig:budget-discovery} summarizes the same LLM runs.}
\label{fig:higher-cost-quality}
\end{figure}
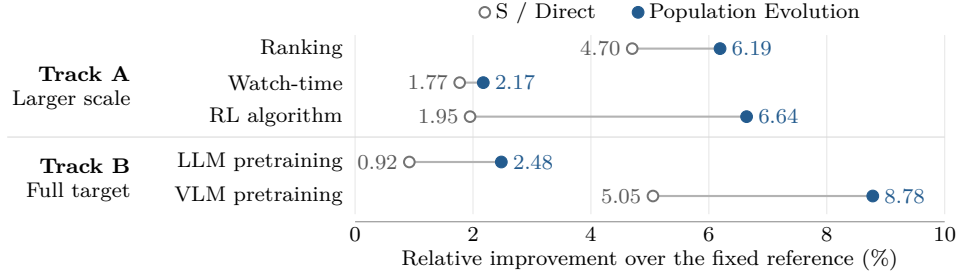

Figure~\ref{fig:higher-cost-quality} reproduces the best larger-scale outcomes from Table~\ref{tab:pe_joint}D for Track~A and the best full-target outcomes from Table~\ref{tab:pe_full} for Track~B.
Each pair uses its domain's metric and fixed reference: Track~A compares PE with S after larger-scale retraining, while Track~B compares PE with Direct after full-target training.
All metrics are oriented so that positive relative gains indicate improvement.

\subsection{Search harnesses and proposal models}
\label{app:harness-proposers}
This comparison tests local discovery and larger-scale transfer across harnesses and proposers.
Table~\ref{tab:pe_compat} holds the watch-time task family and EGMN starting model fixed.
Panel A uses \mbox{GPT-5.5} with AdaEvolve~\citep{cemri2026adaevolve}, EvoX~\citep{liu2026evox}, OpenEvolve~\citep{sharma2025openevolve}, PACEvolve~\citep{yan2026pacevolve}, or AutoResearch~\citep{karpathy2026autoresearch}.
Panel B fixes AdaEvolve and varies the proposer among GPT-5.5, GPT-5.6, Claude Opus 4.8, and Claude Opus 5, all at high reasoning effort.
Each S/PE pair uses the same harness and proposer.
Source columns measure local discovery, while KuaiRand-27K measures larger-scale transfer of the selected designs.

\begin{table}[!ht]
\centering\small
\setlength{\tabcolsep}{3pt}
\renewcommand{\arraystretch}{1.0}
\caption{\textbf{PE improves the best larger-scale gain over S with four of five harnesses and all four proposers.}
Watch-time gains are relative MAE reductions (\%).
Source columns report the best local result on each source instance; KuaiRand-27K reports larger-scale transfer, with passed/tested counts referring to evaluations.
Bold marks the better value within each S/PE pair, using unrounded scores.}
\label{tab:pe_compat}
\begin{tabular*}{\linewidth}{@{\extracolsep{\fill}}llrrrrr@{\hspace{8pt}}rr@{}}
\toprule
& & \multicolumn{5}{c}{Best gain on each source dataset} & \multicolumn{2}{c}{\shortstack{Larger scale\\KuaiRand-27K}} \\
\cmidrule(lr){3-7}\cmidrule(lr){8-9}
Configuration ($k$) & Method & CIKM16 & KuaiRec & WeChat & \shortstack{KuaiRand\\Pure} & \shortstack{KuaiRand\\1K} & \shortstack{Best\\gain} & \shortstack{Passed\\/ tested} \\
\midrule
\multicolumn{9}{@{}l}{\textit{A. Search harnesses (GPT-5.5; high)}} \\
\multirow{2}{*}{AdaEvolve (2)} & S & 7.32 & 0.40 & 0.80 & \textbf{0.75} & 4.93 & 1.77 & 4/10 \\
 & PE & \textbf{12.55} & \textbf{0.67} & \textbf{0.81} & 0.16 & \textbf{5.08} & \textbf{2.17} & 5/7 \\
\addlinespace[2pt]
\multirow{2}{*}{EvoX (4)} & S & 7.01 & 0.28 & \textbf{0.49} & \textbf{0.30} & 1.87 & \textbf{1.45} & 4/20 \\
 & PE & \textbf{11.01} & \textbf{0.58} & 0.49 & 0.11 & \textbf{3.01} & 1.05 & 18/18 \\
\addlinespace[2pt]
\multirow{2}{*}{OpenEvolve (3)} & S & \textbf{7.84} & 0.32 & 0.53 & 0.43 & 4.19 & 1.56 & 12/15 \\
 & PE & 5.72 & \textbf{0.56} & \textbf{0.59} & \textbf{0.59} & \textbf{4.47} & \textbf{1.75} & 11/11 \\
\addlinespace[2pt]
\multirow{2}{*}{PACEvolve (4)} & S & 9.29 & \textbf{0.75} & 0.65 & \textbf{0.95} & \textbf{4.64} & 2.16 & 19/20 \\
 & PE & \textbf{11.14} & 0.46 & \textbf{0.93} & 0.41 & 4.58 & \textbf{2.50} & 20/20 \\
\addlinespace[2pt]
\multirow{2}{*}{AutoResearch (4)} & S & 8.25 & 0.37 & \textbf{0.71} & 0.46 & 3.56 & 1.68 & 20/20 \\
 & PE & \textbf{9.10} & \textbf{0.52} & 0.67 & \textbf{0.73} & \textbf{4.48} & \textbf{2.42} & 16/16 \\
\midrule
\multicolumn{9}{@{}l}{\textit{B. Proposal models (AdaEvolve; high)}} \\
\multirow{2}{*}{GPT-5.5 (2)} & S & 7.32 & 0.40 & 0.80 & \textbf{0.75} & 4.93 & 1.77 & 4/10 \\
 & PE & \textbf{12.55} & \textbf{0.67} & \textbf{0.81} & 0.16 & \textbf{5.08} & \textbf{2.17} & 5/7 \\
\addlinespace[2pt]
\multirow{2}{*}{GPT-5.6 (1)} & S & \textbf{11.52} & 0.55 & \textbf{0.85} & 0.76 & 4.95 & 1.34 & 3/5 \\
 & PE & 11.29 & \textbf{0.56} & 0.69 & \textbf{0.81} & \textbf{5.45} & \textbf{1.50} & 1/3 \\
\addlinespace[2pt]
\multirow{2}{*}{Opus 4.8 (5)} & S & \textbf{12.10} & 0.55 & \textbf{0.96} & 0.78 & 5.00 & 2.61 & 24/25 \\
 & PE & 11.07 & \textbf{0.61} & 0.84 & \textbf{0.91} & \textbf{5.15} & \textbf{2.92} & 20/21 \\
\addlinespace[2pt]
\multirow{2}{*}{Opus 5 (12)} & S & \textbf{15.26} & 1.02 & 1.05 & \textbf{1.63} & 5.42 & 2.61 & 57/60 \\
 & PE & 12.77 & \textbf{1.07} & \textbf{1.28} & 0.78 & \textbf{5.58} & \textbf{3.64} & 48/57 \\
\bottomrule
\end{tabular*}
\par\smallskip
\begin{minipage}{\linewidth}\footnotesize
S: standalone evolution; $k$ is its larger-scale selection quota per source.
Comparisons are within configurations.
AdaEvolve/GPT-5.5 is shared by both panels and Table~\ref{tab:pe_joint}.
Opus denotes Claude Opus.

\end{minipage}
\end{table}

\subsection{LLM full-target quality, yield, and threshold sensitivity}
\label{app:budget-results}
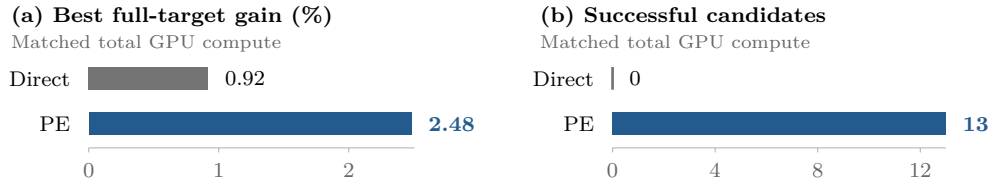
\begin{figure}[!htbp]
\centering
\begin{tikzpicture}[x=1cm,y=1cm,font=\footnotesize]
\node[anchor=west,font=\footnotesize\bfseries] at (0,1.95) {(a) Best full-target gain (\%)};
\node[anchor=west,font=\footnotesize\bfseries] at (7.0,1.95) {(b) Successful candidates};
\node[anchor=west,text=black!60,font=\scriptsize] at (0,1.59) {Matched total GPU compute};
\node[anchor=west,text=black!60,font=\scriptsize] at (7.0,1.59) {Matched total GPU compute};
\foreach \y/\method in {1.13/Direct,0.53/PE} {
  \node[anchor=east] at (1.03,\y) {\method};
  \node[anchor=east] at (7.96,\y) {\method};
}
\fill[black!55] (1.15,.98) rectangle (2.73,1.28);
\fill[peblue] (1.15,.38) rectangle (5.42,.68);
\node[anchor=west] at (2.83,1.13) {0.92};
\node[anchor=west,font=\footnotesize\bfseries,text=peblue] at (5.52,.53) {2.48};
\draw[black!30] (1.15,.21) -- (5.46,.21);
\foreach \x/\v in {1.15/0,2.87/1,4.59/2} {
  \draw[black!30] (\x,.21) -- (\x,.15);
  \node[anchor=north,text=black!60] at (\x,.13) {\v};
}
\draw[black!55,line width=1pt] (8.08,.98) -- (8.08,1.28);
\fill[peblue] (8.08,.38) rectangle (12.49,.68);
\node[anchor=west] at (8.18,1.13) {0};
\node[anchor=west,font=\footnotesize\bfseries,text=peblue] at (12.59,.53) {13};
\draw[black!30] (8.08,.21) -- (12.49,.21);
\foreach \x/\v in {8.08/0,9.437/4,10.794/8,12.151/12} {
  \draw[black!30] (\x,.21) -- (\x,.15);
  \node[anchor=north,text=black!60] at (\x,.13) {\v};
}
\end{tikzpicture}
\caption{\textbf{PE improves LLM quality and discovery yield under matched total GPU compute.}
Both panels summarize completed runs.
(a) Best relative full-target gains are 2.48\% for PE and 0.92\% for Direct.
(b) The predefined 1\% criterion is met by 13 candidates from 39 PE full-target attempts and none from 70 Direct attempts.}
\label{fig:budget-discovery}
\label{fig:useful-diversity}
\end{figure}

Figure~\ref{fig:budget-discovery} summarizes the completed LLM discovery runs in Section~\ref{sec:full-results}, which use matched total GPU compute.
Panel (a) reports the best full-target gain; panel (b) reports the number of candidates meeting the predefined 1\% relative-improvement criterion.
All executed search and evaluation stages contribute to cost under the accounting in Appendix~\ref{app:execution-accounting}.

\paragraph{Sensitivity to the success threshold.}
We reclassify the completed full-target results at four relative-improvement thresholds while retaining the fixed reference accuracy of 40.25\% (0.4025 on the unit scale).
A candidate meets threshold $\tau$ when its score is at least $0.4025(1+\tau)$.
The main threshold remains the 1\% criterion fixed before the reported comparisons.
Table~\ref{tab:llm-threshold-sensitivity} shows that PE returns more successful candidates at every examined threshold.
At 0.5\%, Direct returns six successful candidates and PE returns 21; at 1\%, 1.5\%, and 2\%, Direct returns none while PE returns 13, eight, and three, respectively.
The best Direct gain is 0.92\%, while the best PE gain is 2.48\%; the quality comparison complements the threshold-dependent success counts.

\begin{table}[!htbp]
\centering\small
\setlength{\tabcolsep}{10pt}
\renewcommand{\arraystretch}{1.1}
\caption{\textbf{LLM success counts across full-target gain thresholds.} Counts use the same completed runs under matched total GPU compute; the predefined main threshold is bold.}
\label{tab:llm-threshold-sensitivity}
\begin{tabular}{@{}lrr@{}}
\toprule
Relative gain threshold (\%) & PE & Direct \\
\midrule
0.50 & 21 & 6 \\
\textbf{1.00} & \textbf{13} & \textbf{0} \\
1.50 & 8 & 0 \\
2.00 & 3 & 0 \\
\bottomrule
\end{tabular}
\end{table}
\FloatBarrier

\section{Transfer to New Datasets and Coding Tasks}
\label{app:postdiscovery-transfer}

\subsection{Candidate selection and evaluation protocol}
\label{app:transfer-protocol}
This appendix specifies the three transfer evaluations in Section~\ref{sec:benchmark-generalization}, testing the designs returned by each complete discovery procedure beyond its discovery instances.
Watch-time heads transfer to VK-LSVD; ranking designs transfer to KuaiVideo with a change from full-catalog next-item prediction to candidate-conditioned click prediction; RL modifications are evaluated on a task partition disjoint from the 423-task larger-scale selection set.
The two recommendation datasets were excluded from source search, peer evaluation, and larger-scale promotion.

\paragraph{Candidate selection and evaluation.}
Each procedure supplies candidates through its own selection rule. Each discovered intervention is evaluated without candidate-specific redesign or hyperparameter tuning, using a shared target protocol.
For watch-time, PE contributes designs that passed larger-scale evaluation, and S contributes the best source candidate from each of five datasets.
For ranking, PE likewise contributes designs that passed larger-scale evaluation, and S contributes the top four source candidates from each of five datasets.
RL compares two successful PE candidates and the best S candidate from each of five source searches.
Recommendation weights are trained from scratch on each new dataset; RL policies start from the same pretrained model and follow a fixed training protocol before evaluation on held-out coding tasks.

\paragraph{Target calibration and reporting.}
Each recommendation target uses one baseline-calibrated configuration shared by the reference and every candidate, including the training duration.
Baseline hyperparameter and training-duration selection use the released test split, which is also used for final reporting; no separate validation split is used for these selection decisions.
The recommendation results therefore measure transfer to new datasets under shared target-side calibration, rather than an independent test evaluation after calibration.
Candidate-specific HPO and evolutionary adaptation on these targets are disabled.
The RL reference uses the fixed, unmodified training configuration without target-side HPO; the 632 coding tasks are disjoint from the larger-scale selection partition.
Recommendation references and candidates are trained with three seeds; RL references and candidates undergo three independent stochastic reruns with the same nominal training seed.
We report means and sample standard deviations over these three runs. These repetitions assess candidate performance rather than variability across complete discovery runs.
AUC, gAUC, and pass@1 are expressed as percentages; MAE is in seconds. Means and standard deviations use two decimal places, except log loss, which uses four. All comparisons use the original precision.

\paragraph{Candidate-set and best-candidate summaries.}
For recommendation, each candidate's score is its three-seed mean, and candidate-set means give each evaluated candidate record equal weight.
Candidate-set means measure average transfer of each procedure's selected designs; best-candidate scores identify its strongest evaluated design by mean AUC or mean MAE.
On VK-LSVD, PE and S obtain $14.54\pm0.01$ s and $14.55\pm0.01$ s. For each seed, we first average MAE across candidates; the reported standard deviation is then computed across these three averages.
On KuaiVideo, the corresponding mean AUCs are 74.45\% and 74.39\%.
As a weighting check, we group behaviorally equivalent implementations and aggregate their unrounded scores; equivalence is determined from implementations, not equality of rounded metrics. Counting each group once preserves the PE--S ordering: mean AUCs remain 74.45\% and 74.39\%, and mean MAEs remain $14.54\pm0.01$ s and $14.55\pm0.01$ s after rounding.
The five RL source searches share the same training dataset and objective, and their discovery goal is to find the strongest algorithm for that setting.
We therefore compare the best RL candidate from each method using its mean over three stochastic reruns, rather than the candidate-set mean.
Table~\ref{tab:transfer-generalization} summarizes the PE--S comparisons; the following subsections report the references and individual candidates.

\subsection{Watch-time transfer to VK-LSVD}
\label{app:transfer-watchtime}

This evaluation tests whether a discovered distributional head remains useful on a new dataset under a common feature encoder and training protocol.

\paragraph{Data and shared model.}
We use the HEGM-processed {\small\texttt{VK-LSVD/up0.01\_ir0.01}} subset of VK-LSVD~\citep{poslavsky2026vklsvd,gulevskaia2026hegm}, retaining its chronological split and precomputed features.\footnote{Raw data: \url{https://huggingface.co/datasets/deepvk/VK-LSVD}; processed release: \url{https://github.com/rw404/HEGM}.}
Video duration and the remaining numerical features are retained; duration buckets and categorical vocabularies are constructed from training data only.
All candidates share the encoder and training settings in Table~\ref{tab:vk-transfer-protocol}; only the distributional head changes.

\begin{table}[htbp]
\centering\small
\setlength{\tabcolsep}{5pt}
\renewcommand{\arraystretch}{1.1}
\caption{\textbf{Shared VK-LSVD transfer protocol.} Final candidates train on the released training split and are evaluated on the released test split.}
\label{tab:vk-transfer-protocol}
\begin{tabular}{@{}P{3.3cm}P{9.2cm}@{}}
\toprule
Component & Specification \\
\midrule
Examples (train / val. / test) & 7,505,820 / 1,516,831 / 1,275,643 \\
Chronology & 5/1/1-week split after historical-feature weeks 0--19; training uses weeks 20--24 \\
Numerical features & Video duration and 31 other features; normalization constant: 157 seconds \\
Categorical features & Age, gender, geographic region, place, platform, and user agent \\
Duration encoding & 50 quantile-based buckets fitted on training data \\
Shared EGMN encoder & 16-dimensional categorical embeddings, scalar linear projections, and a 256/128/64 MLP \\
Reference head & One exponential and ten left-truncated Gaussian components \\
Reference loss & Mixture likelihood, L1 point-prediction error, and entropy regularization \\
Training & Adagrad; batch size 2,048; ten epochs; three training seeds \\
\bottomrule
\end{tabular}
\end{table}

\paragraph{Calibration and outcomes.}
Baseline calibration monitors released-test MAE and completes 11 valid trials within a cap of 50, stopping after ten successive trials without improvement.
Table~\ref{tab:heldout-hpo-config} lists the search ranges and selected values shared by all candidates; no candidate-specific tuning follows.
The shared calibrated EGMN reference has MAE $14.73\pm0.02$ s. Every evaluated PE head improves its mean MAE, with paired reductions of 1.02--1.36\% (Table~\ref{tab:watchtime-candidate-results}). All evaluated S heads also improve over this reference.
PE has lower candidate-set mean MAE (Table~\ref{tab:transfer-generalization}), although the strongest individual head comes from S: $14.46\pm0.02$ s versus $14.53\pm0.02$ s for PE.
The WeChat iteration-21 head achieves $14.58\pm0.03$ s and a $(1.02\pm0.30)\%$ paired MAE reduction, alongside its source-instance and KuaiRand-27K improvements.
This is a concrete example of one unchanged distributional design remaining useful across datasets and training scales.

\begin{table}[!htbp]
\centering\small
\setlength{\tabcolsep}{6pt}
\renewcommand{\arraystretch}{1.1}
\caption{\textbf{VK-LSVD reference and all evaluated transfer candidates.} Values are means $\pm$ sample standard deviations across three training seeds. Reductions are computed against the seed-matched reference. PE identifiers are archived candidate-ID suffixes; S entries identify their source dataset.}
\label{tab:watchtime-candidate-results}
\begin{tabular*}{\linewidth}{@{\extracolsep{\fill}}llrr@{}}
\toprule
Method & Candidate / source & MAE (s) $\downarrow$ & Paired reduction (\%) $\uparrow$ \\
\midrule
Reference & EGMN & $14.73\pm0.02$ & --- \\
\midrule
PE & \texttt{e33792c7}$^{\dagger}$ & $14.53\pm0.02$ & $1.36\pm0.15$ \\
PE & \texttt{2f3242e3}$^{\dagger}$ & $14.53\pm0.02$ & $1.36\pm0.15$ \\
PE & \texttt{92f5eb20}$^{\dagger}$ & $14.53\pm0.02$ & $1.36\pm0.15$ \\
PE & \texttt{e0b2ffa7} & $14.53\pm0.01$ & $1.35\pm0.06$ \\
PE & \texttt{94db2d04} & $14.58\pm0.03$ & $1.02\pm0.30$ \\
\midrule
S & KuaiRand-1K & $14.46\pm0.02$ & $1.80\pm0.10$ \\
S & WeChat & $14.51\pm0.01$ & $1.47\pm0.15$ \\
S & KuaiRand-Pure & $14.55\pm0.00$ & $1.23\pm0.12$ \\
S & CIKM16 & $14.56\pm0.04$ & $1.12\pm0.39$ \\
S & KuaiRec & $14.69\pm0.01$ & $0.24\pm0.17$ \\
\bottomrule
\end{tabular*}
\par\smallskip
\begin{minipage}{\linewidth}\footnotesize
$^{\dagger}$These candidates are behaviorally equivalent implementations. Aggregation uses the original, unrounded run scores.
\end{minipage}
\end{table}
\FloatBarrier

\subsection{Ranking transfer to KuaiVideo click prediction}
\label{app:transfer-ranking}

This evaluation tests sequence-model designs under both a new dataset and a different recommendation objective.
A common adapter makes the starting model and selected designs compatible with candidate-conditioned click prediction.

\paragraph{Data and features.}
We use the RecZoo \texttt{KuaiVideo\_x1} release with its original train/test split and ALPINE-processed visual features~\citep{li2019alpine}.\footnote{Release and feature archive: \url{https://huggingface.co/datasets/reczoo/KuaiVideo_x1}; visual features: \texttt{item\_visual\_emb\_dim64.h5}.}
Each example contains a user, candidate item, positive and negative histories, and a binary click label; histories retain the released order and are truncated to the most recent 100 items.
Categorical vocabularies use training data only, while the visual feature table remains frozen without additional normalization or dimensionality reduction.
The visual feature table provides complete coverage, including the 41.81\% of test rows whose candidate items map to the categorical unknown token. Users are represented by learned categorical embeddings; the released user-visual features are not used.
Table~\ref{tab:kuai-transfer-protocol} summarizes the data and shared adapter specifications.

\begin{table}[htbp]
\centering\small
\setlength{\tabcolsep}{5pt}
\renewcommand{\arraystretch}{1.1}
\caption{\textbf{Shared KuaiVideo data and adapter specifications.} The discovered sequence-model code is the only component that differs among the reference, S, and PE candidates.}
\label{tab:kuai-transfer-protocol}
\begin{tabular}{@{}P{3.5cm}P{9.0cm}@{}}
\toprule
Component & Specification \\
\midrule
Candidate rows (train / test) & 10,931,092 / 2,730,291 \\
Requests (train / test) & 803,007 / 209,728; 10,000 users per split \\
Item vocabulary & Minimum training count 10; 632,404 known items, plus padding and unknown entries \\
Visual features & 64-dimensional vectors for 3,242,315 items; frozen, with separate padding and unknown entries \\
Item representation & Learned item-ID embedding plus a trainable, bias-free $64\!\rightarrow\!128$ projection of its visual vector \\
History encoding & Shared sequence encoder and embeddings for positive and negative histories; at most 100 items each \\
Click predictor & MLP widths 512/256/1; binary cross-entropy loss \\
Final training & Four epochs; four H200 GPUs; three training seeds \\
\bottomrule
\end{tabular}
\end{table}

\paragraph{Common sequence-aware CTR adapter.}
The candidate item is appended as the final query token to each history. Its output states $h^+$ and $h^-$ summarize the positive and negative histories conditioned on that item.
Because these histories lack per-item timestamps, the adapter supplies sequential positions $1,\ldots,L+1$.
Given candidate embedding $e$ and user embedding $u$, the shared click predictor uses

\begin{equation*}
\hat p=\sigma\!\left(\operatorname{MLP}\!\left([h^+;h^-;e;h^+-h^-;h^+\odot e;h^-\odot e;u]\right)\right).
\end{equation*}
The reference, S candidates, and PE candidates use this identical adapter and differ only in the discovered sequence-model design.

\paragraph{Calibration and final training.}
Baseline calibration uses 50 serial Optuna TPE trials, permitting two to five epochs with early-stopping patience two. Hyperparameters and the training duration are selected using released-test AUC.
The selected configuration in Table~\ref{tab:heldout-hpo-config} is shared by all candidates, which train for exactly four epochs and are evaluated once afterward.
Each candidate is trained under three seeds; AUC is the primary metric, with sample-weighted user gAUC and binary log loss as auxiliary metrics.

\begin{table}[htbp]
\centering\small
\setlength{\tabcolsep}{5pt}
\renewcommand{\arraystretch}{1.08}
\caption{\textbf{Shared calibration for the new recommendation targets.} Baseline configuration and training-duration selection use the released test split; the selected protocol is reused unchanged for the reference and every candidate.}
\label{tab:heldout-hpo-config}
\begin{tabular*}{\linewidth}{@{\extracolsep{\fill}}lll@{}}
\toprule
Parameter & Search range & Shared selected value \\
\midrule
\multicolumn{3}{l}{\textit{VK-LSVD}} \\
Learning rate & $[0.01,0.3]$ & 0.076611 \\
Weight decay & $[10^{-7},10^{-3}]$ & $4.20799\times10^{-7}$ \\
Entropy coefficient $\alpha$ & $[0,1]$ & 0.37454 \\
Regression coefficient $\beta$ & $[0.1,5]$ & 4.7585 \\
Dropout & $[0,0.5]$ & 0.365997 \\
\midrule
\multicolumn{3}{l}{\textit{KuaiVideo}} \\
Learning rate & Log-uniform $[10^{-4},6\times10^{-3}]$ & $2.9416\times10^{-4}$ \\
Weight decay & Log-uniform $[10^{-6},10^{-3}]$ & $5.3419\times10^{-6}$ \\
Dropout & $\{0.1,0.2,0.3,0.5\}$ & 0.1 \\
Global batch size & $\{1024,2048\}$ & 2,048 \\
Epochs & 2--5 (patience 2) & 4 \\
\bottomrule
\end{tabular*}
\end{table}
\begin{table}[htbp]
\centering\small
\setlength{\tabcolsep}{3pt}
\renewcommand{\arraystretch}{1.12}
\caption{\textbf{KuaiVideo reference and best S and PE designs.}
The best S and PE designs are selected by mean AUC across three training seeds; all metrics in a row come from that same design.
Every metric is reported as mean $\pm$ sample standard deviation over those same runs.}
\label{tab:transfer-ranking-metrics}
\begin{tabular*}{\linewidth}{@{\extracolsep{\fill}}lrrr@{}}
\toprule
Model & AUC (\%) $\uparrow$ & User gAUC (\%) $\uparrow$ & Log loss $\downarrow$ \\
\midrule
Reference & $74.39\pm0.11$ & $66.14\pm0.06$ & $0.4378\pm0.0008$ \\
Best S & $74.43\pm0.08$ & $66.27\pm0.05$ & $0.4371\pm0.0018$ \\
Best PE & $\mathbf{74.46\pm0.06}$ & $\mathbf{66.29\pm0.02}$ & $\mathbf{0.4365\pm0.0005}$ \\
\bottomrule
\end{tabular*}
\end{table}

\paragraph{Transfer outcome.}
The best PE candidate, discovered on KuaiRand-Pure, adds a gated spectral causal long-filter channel.
Its mean AUC exceeds that of the best S design from KuaiRand-1K by 0.04 percentage points computed from unrounded means, with improvements in both auxiliary metrics (Table~\ref{tab:transfer-ranking-metrics}).
Every evaluated PE design exceeds the shared calibrated reference in mean AUC, with increases of approximately 0.04--0.07 percentage points computed before rounding (Table~\ref{tab:ranking-pe-candidates}). Eight evaluated S candidates exceed the reference mean (Table~\ref{tab:ranking-s-candidates}). At the candidate-set level, mean AUC is 74.45\% for PE and 74.39\% for S.
These results support design transfer under a common CTR adapter and shared target-side calibration; the small PE--S differences are descriptive comparisons of three-seed means.

\begin{table}[!htbp]
\centering\footnotesize
\setlength{\tabcolsep}{4pt}
\renewcommand{\arraystretch}{1.08}
\caption{\textbf{All evaluated PE ranking candidates on KuaiVideo.} Identifiers are archived candidate-ID suffixes. Values are means $\pm$ sample standard deviations across three training seeds. Every candidate's mean AUC exceeds the reference before rounding.}
\label{tab:ranking-pe-candidates}
\begin{tabular*}{\linewidth}{@{\extracolsep{\fill}}lrrr@{}}
\toprule
Candidate & AUC (\%) $\uparrow$ & User gAUC (\%) $\uparrow$ & Log loss $\downarrow$ \\
\midrule
Reference & $74.39\pm0.11$ & $66.14\pm0.06$ & $0.4378\pm0.0008$ \\
\midrule
\texttt{b3cb61d8} & $74.46\pm0.06$ & $66.29\pm0.02$ & $0.4365\pm0.0005$ \\
\texttt{5f007905} & $74.46\pm0.07$ & $66.26\pm0.01$ & $0.4371\pm0.0006$ \\
\texttt{7ac9ec81} & $74.46\pm0.09$ & $66.29\pm0.09$ & $0.4369\pm0.0011$ \\
\texttt{e7ec2459} & $74.46\pm0.09$ & $66.29\pm0.09$ & $0.4369\pm0.0011$ \\
\texttt{2755e3d1} & $74.46\pm0.09$ & $66.29\pm0.09$ & $0.4369\pm0.0011$ \\
\texttt{1add4a4b} & $74.46\pm0.09$ & $66.29\pm0.09$ & $0.4369\pm0.0011$ \\
\texttt{8582c2be} & $74.45\pm0.08$ & $66.27\pm0.04$ & $0.4377\pm0.0007$ \\
\texttt{d54779e3} & $74.45\pm0.08$ & $66.27\pm0.04$ & $0.4377\pm0.0007$ \\
\texttt{33963c64} & $74.45\pm0.08$ & $66.27\pm0.05$ & $0.4377\pm0.0008$ \\
\texttt{314d2b6b} & $74.45\pm0.11$ & $66.27\pm0.10$ & $0.4370\pm0.0010$ \\
\texttt{4b146074} & $74.45\pm0.11$ & $66.27\pm0.10$ & $0.4370\pm0.0010$ \\
\texttt{c9913f8e} & $74.43\pm0.09$ & $66.26\pm0.04$ & $0.4377\pm0.0010$ \\
\texttt{ad9fa175} & $74.43\pm0.09$ & $66.26\pm0.04$ & $0.4377\pm0.0009$ \\
\texttt{fff8c656} & $74.43\pm0.08$ & $66.26\pm0.07$ & $0.4375\pm0.0006$ \\
\texttt{17f754bc} & $74.43\pm0.08$ & $66.26\pm0.03$ & $0.4377\pm0.0009$ \\
\texttt{a3da30b7} & $74.43\pm0.09$ & $66.26\pm0.04$ & $0.4377\pm0.0010$ \\
\texttt{648f09a3} & $74.43\pm0.09$ & $66.26\pm0.03$ & $0.4378\pm0.0010$ \\
\texttt{fd6457ef} & $74.43\pm0.09$ & $66.25\pm0.03$ & $0.4377\pm0.0009$ \\
\bottomrule
\end{tabular*}
\end{table}

\begin{table}[!htbp]
\centering\footnotesize
\setlength{\tabcolsep}{3pt}
\renewcommand{\arraystretch}{1.08}
\caption{\textbf{All evaluated S ranking candidates on KuaiVideo.} Each source contributes its top four candidates by source performance. Identifiers are archived candidate-ID suffixes; values are means $\pm$ sample standard deviations across three training seeds.}
\label{tab:ranking-s-candidates}
\begin{tabular*}{\linewidth}{@{\extracolsep{\fill}}llrrr@{}}
\toprule
Source & Candidate & AUC (\%) $\uparrow$ & User gAUC (\%) $\uparrow$ & Log loss $\downarrow$ \\
\midrule
--- & Reference & $74.39\pm0.11$ & $66.14\pm0.06$ & $0.4378\pm0.0008$ \\
\midrule
KuaiRand-1K & \texttt{560d182a} & $74.43\pm0.08$ & $66.27\pm0.05$ & $0.4371\pm0.0018$ \\
KuaiRand-1K & \texttt{35335eb3} & $74.42\pm0.08$ & $66.27\pm0.03$ & $0.4371\pm0.0013$ \\
KuaiRand-1K & \texttt{e5c01c96} & $74.42\pm0.08$ & $66.27\pm0.03$ & $0.4371\pm0.0013$ \\
KuaiRand-1K & \texttt{5d08ec24} & $74.42\pm0.08$ & $66.26\pm0.04$ & $0.4371\pm0.0013$ \\
KuaiSAR & \texttt{3a70637c} & $74.41\pm0.10$ & $66.28\pm0.07$ & $0.4368\pm0.0014$ \\
KuaiSAR & \texttt{6aa0c6a0} & $74.41\pm0.09$ & $66.28\pm0.07$ & $0.4368\pm0.0013$ \\
KuaiSAR & \texttt{c576350a} & $74.40\pm0.09$ & $66.28\pm0.07$ & $0.4368\pm0.0013$ \\
KuaiSAR & \texttt{a3707be1} & $74.40\pm0.08$ & $66.27\pm0.06$ & $0.4369\pm0.0009$ \\
KuaiRand-Pure & \texttt{fb4fcfa1} & $74.38\pm0.04$ & $66.26\pm0.03$ & $0.4372\pm0.0001$ \\
KuaiRand-Pure & \texttt{e520862d} & $74.38\pm0.04$ & $66.26\pm0.03$ & $0.4373\pm0.0002$ \\
KuaiRand-Pure & \texttt{63cd4a7d} & $74.38\pm0.04$ & $66.26\pm0.03$ & $0.4372\pm0.0001$ \\
KuaiRand-Pure & \texttt{3a338551} & $74.37\pm0.04$ & $66.26\pm0.03$ & $0.4372\pm0.0001$ \\
KuaiLive & \texttt{2577bcf4} & $74.37\pm0.11$ & $66.23\pm0.04$ & $0.4375\pm0.0009$ \\
KuaiLive & \texttt{8b954d65} & $74.37\pm0.11$ & $66.23\pm0.04$ & $0.4375\pm0.0009$ \\
KuaiRec & \texttt{7d3732b5} & $74.37\pm0.07$ & $66.20\pm0.02$ & $0.4385\pm0.0005$ \\
KuaiRec & \texttt{384908be} & $74.37\pm0.07$ & $66.20\pm0.02$ & $0.4385\pm0.0005$ \\
KuaiLive & \texttt{d895ff6d} & $74.37\pm0.11$ & $66.23\pm0.04$ & $0.4374\pm0.0009$ \\
KuaiRec & \texttt{be23d8b2} & $74.36\pm0.06$ & $66.20\pm0.02$ & $0.4385\pm0.0004$ \\
KuaiRec & \texttt{4b23990b} & $74.36\pm0.06$ & $66.20\pm0.02$ & $0.4385\pm0.0005$ \\
KuaiLive & \texttt{9e6f90f4} & $74.36\pm0.12$ & $66.23\pm0.04$ & $0.4375\pm0.0009$ \\
\bottomrule
\end{tabular*}
\end{table}
\FloatBarrier

\subsection{RL transfer to a disjoint LiveCodeBench partition}
\label{app:transfer-rl}

This evaluation tests policies trained with the discovered RL algorithms on held-out coding tasks under a fixed training protocol.

\paragraph{Task partition and training.}
The 1,055 LiveCodeBench tasks comprise 423 larger-scale selection tasks and 632 held-out tasks with disjoint IDs, both scored using private tests.
All methods use Qwen2.5-Coder-3B-Instruct, the same larger-scale configuration, and 50 GRPO rollout steps.
Candidates change the GRPO loss and/or rollout filtering; held-out coding tasks are used only to evaluate the trained policies.
Each candidate is freshly trained and evaluated in three stochastic reruns with the same nominal training seed.

\paragraph{Best-candidate comparison.}
PE obtains mean pass@1 of $(23.10\pm0.32)\%$, compared with $(22.94\pm0.16)\%$ for S (Table~\ref{tab:transfer-generalization}).
The unmodified reference reaches $(23.26\pm0.42)\%$ pass@1. The reference, best PE candidate, and best S candidate solve 147, 146, and 145 tasks on average, respectively (Table~\ref{tab:rl-transfer-candidates}). Thus PE exceeds S by one solved task on average, while neither candidate exceeds the reference mean. These are descriptive comparisons across three stochastic reruns.

\begin{table}[!htbp]
\centering\small
\setlength{\tabcolsep}{6pt}
\renewcommand{\arraystretch}{1.1}
\caption{\textbf{Reference and all evaluated RL candidates on LiveCodeBench-632.} Each entry uses three independent stochastic reruns with nominal training seed 42. Pass@1 is reported as mean $\pm$ sample standard deviation. The reference uses the unmodified configuration without target-side HPO.}
\label{tab:rl-transfer-candidates}
\begin{tabular*}{\linewidth}{@{\extracolsep{\fill}}llcr@{}}
\toprule
Method & Candidate & Solved / 632 in each rerun & Pass@1 (\%) $\uparrow$ \\
\midrule
Reference & --- & 145, 150, 146 & $23.26\pm0.42$ \\
\midrule
PE & \texttt{93d95a15} & 146, 143, 139 & $22.57\pm0.56$ \\
PE & \texttt{cbf1c653} & 146, 144, 148 & $23.10\pm0.32$ \\
\midrule
S & \texttt{a0936bce} & 141, 143, 142 & $22.47\pm0.16$ \\
S & \texttt{db21a301} & 146, 145, 144 & $22.94\pm0.16$ \\
S & \texttt{77431a4e} & 136, 140, 136 & $21.73\pm0.37$ \\
S & \texttt{86561efd} & 140, 143, 149 & $22.78\pm0.73$ \\
S & \texttt{754c241c} & 131, 130, 138 & $21.04\pm0.69$ \\
\bottomrule
\end{tabular*}
\end{table}
\FloatBarrier

\section{Limitations and Future Directions}
\label{app:limitations}

RMD-Bench covers five task families, with compatible intervention interfaces within each family.
The transfer results apply to the architectures, training protocols, and evaluation settings studied here; extending PE to less compatible tasks may require additional interface design.
Track~A evaluates design generalization, whereas Track~B targets improvement on a specified expensive workload.

The source-level comparison tests shared experimental feedback; larger-scale and full-target results evaluate the complete discovery procedures.
Given the cost of full runs, further independent searches would help characterize variability in discovered designs and final performance (Appendix~\ref{app:execution-accounting}).
Future work can also examine alternative exchange and promotion policies and extend the benchmark to additional task families.

\end{document}